\documentclass{article}

\usepackage{microtype}
\usepackage{graphicx}
\usepackage{subfigure}
\usepackage{float}
\usepackage{booktabs} 

\usepackage{hyperref}

\usepackage[accepted]{icml2023}

\usepackage{amsmath}
\usepackage{amssymb}
\usepackage{mathtools}
\usepackage{amsthm}

\usepackage[capitalize,noabbrev]{cleveref}

\usepackage{multirow}
\usepackage{stfloats}
\usepackage{dutchcal}

\theoremstyle{plain}
\newtheorem{theorem}{Theorem}[section]

\newtheorem{lemma}[theorem]{Lemma}

\theoremstyle{definition}

\newtheorem{assumption}{Assumption}
\newtheorem{remark}[theorem]{Remark}

\usepackage[textsize=tiny]{todonotes}

\icmltitlerunning{Dynamic Regularized Sharpness Aware Minimization in Federated Learning}

\begin{document}

\twocolumn[
\icmltitle{Dynamic Regularized Sharpness Aware Minimization in Federated Learning: 
\\Approaching Global Consistency and Smooth Landscape
}




\begin{icmlauthorlist}
\icmlauthor{Yan Sun}{sch}
\icmlauthor{Li Shen}{comp}
\icmlauthor{Shixiang Chen}{comp}
\icmlauthor{Liang Ding}{comp}
\icmlauthor{Dacheng Tao}{sch}
\end{icmlauthorlist}

\icmlaffiliation{sch}{The University of Sydney, Institute of Engineering, Sydney NSW, Australia}
\icmlaffiliation{comp}{JD Explore Academy, Beijing, China}

\icmlcorrespondingauthor{Li Shen}{mathshenli@gmail.com}

\icmlkeywords{Machine Learning, ICML}

\vskip 0.3in
]



\printAffiliationsAndNotice{} 

\begin{abstract}
\label{abstract}
In federated learning (FL), a cluster of local clients are chaired under the
coordination of the global server and cooperatively train one model with privacy protection. 
Due to the multiple local updates and the isolated non-iid dataset, clients are prone to overfit into their own optima, which extremely deviates from the global objective and significantly undermines the performance.
Most previous works only focus on enhancing the consistency between the local and global objectives to alleviate this prejudicial client drifts from the perspective of the optimization view, whose performance would be prominently deteriorated on the high heterogeneity.
In this work, we propose a novel and general algorithm {\ttfamily FedSMOO} by jointly considering the optimization and generalization targets to efficiently improve the performance in FL.
Concretely, {\ttfamily FedSMOO} adopts a dynamic regularizer to guarantee the local optima towards the global objective, which is meanwhile revised by the global Sharpness Aware Minimization (SAM) optimizer to search for the consistent flat minima.
Our theoretical analysis indicates that {\ttfamily FedSMOO} achieves fast $\mathcal{O}(1/T)$ convergence rate with low generalization bound.
Extensive numerical studies are conducted on the real-world dataset to verify its peerless efficiency and excellent generality.
Our code is available at \url{https://github.com/woodenchild95/FL-Simulator.git}.
\end{abstract}
\section{Introduction}
\label{introduction}

FL is a horizontally distributed framework that allows a mass of edge devices to collaboratively train a global model on their isolated private dataset~\cite{FedAvg}. To protect the data privatization and localization, each client has no direct access to the other dataset. To enable information exchange during training, a global server is employed to communicate the intermediate variables and parameters with the local clients, and to assist them with performing the joint training. Due to the poor bandwidth throttling especially on the global server, it adopts multiple local training and partial participation to mitigate the communication bottleneck to a greater extent. With extensive studies of FL, theoretical analysis reveals that the major influence on limiting the performance of FL is client drifts, whose essence is that the inconsistent local optima deviate from the global objective on the heterogeneous dataset~\cite{SCAFFOLD,local_sgd,develop1,develop2}. \citet{linear_speedup} theoretically demonstrate that the performance of the classical {\ttfamily FedAvg} method suffers from the length of local updates and the number of partial participation multiplied by the constant upper bound of the variance of the heterogeneous gradient, which contributes as the dominant term of the convergence rate. This divergence would be extremely multiplied by both increasing the local interval and reducing the participation ratio.

To address these problems above, spontaneously, most existing works focus on tackling the global consistent objectives from the perspective of optimization view with Empirical Risk Minimization (ERM)~\cite{inconsistency2,inconsistency3,inconsistency1, consistency,zhang2022fine}. With the process of global training, steady correction to the local objectives based on the regularization term productively reduces the upper bound of theoretical convergence rate~\cite{FedPD, FedDyn}. Furthermore, considering the asynchrony, re-weighted empirical averages contribute to aggregating more stable global parameters~\cite{FedNova,huang2022stochastic}. However, in the face of the challenge of a higher heterogeneous dataset, e.g. for heavy tail, the global solution falling into a sharp landscape may fail to provide solid estimates and yet overfit to an unreliable minimum with poor stability~\cite{heavy_tail, SAM}. Therefore, we need to contemplate the training process comprehensively and completely. More attention must be paid to the generalization ability under ensuring the optimization performance in the federated framework.

To further approach the consistent stationary state with credible stability in FL, in this paper, we jointly take into account ensuring both the optimization and generalization performance on the global target and propose a novel algorithm {\ttfamily FedSMOO} to guide the local clients to search for the consistent flat minima aligned to the global objective which significantly improve the performance, especially on the dataset with high heterogeneity. Specifically, we first adopt the dynamic regularizer to align the global and local objectives. Under the assurance of local consistency, we consider the global objective as searching for a flat stationary state based on the global SAM optimizer, which will correct the regularization term dynamically. As the local multiple updates, applying the general SAM optimizer on the global server cannot get the accurate perturbation when updating locally. To tackle this, we introduce an additional equality constraint on the perturbation in the global optimization objective and employ another regularization term to modify the updates of SAM, which maintains the consistency of the local perturbation terms aligned to the global target. When the active local clients converge, they eventually achieve the global stationary state with a smooth loss landscape.

In contrast with \citet{FedSAM}~({\ttfamily FedSAM}) which firstly incorporates the SAM optimizer with FL and allows each client to train the model with a local SAM optimizer, they focus on improving the \textit{local generality} to inherently align the local flatness closer to the global flatness and expect to enhance the consistency. {\ttfamily FedSMOO} jointly considers the global consistency and \textit{global generality} as the target from the perspective of both the optimization and generalization to approach a consistent smooth minimum.

Theoretically, we relax the general assumption of bounded heterogeneous gradients and demonstrate that our proposed {\ttfamily FedSMOO} could achieve the fast $\mathcal{O}(1/T)$ convergence rate on smooth non-convex scenes. Then we provide the generalization bound of the global function to theoretically guarantee the smoothness of the optimal state. Empirically, we conduct extensive experiments on the CIFAR-10$/$100 dataset and show that {\ttfamily FedSMOO} achieves the faster convergence rate and higher generalization accuracy in practice, compared with $7$ baselines including {\ttfamily FedAvg}, {\ttfamily FedAdam}, {\ttfamily SCAFFOLD}, {\ttfamily FedCM}, {\ttfamily FedDyn}, {\ttfamily FedSAM}, and {\ttfamily MoFedSAM}.

In the end, we summarize our main contributions as:
\begin{itemize}
    \item We propose a novel and general federated algorithm, {\ttfamily FedSMOO}, to jointly consider both global consistency and global generality as the target, which simultaneously stays fast convergence and high generalization.
    \item We provide the theoretical analysis of the convergence rate and generalization bound. On the smooth non-convex scene, {\ttfamily FedSMOO} could achieve the fast $\mathcal{O}(1/T)$ convergence rate without the general assumption of the bounded heterogeneous gradients.
    \item Extensive numerical studies are conducted on the CIFAR-10$/$100 dataset to verify the excellent performance of {\ttfamily FedSMOO}, which outperforms several classical baselines, especially on the high heterogeneity.
\end{itemize}

\section{Related Work}
\label{related work}
\textbf{FL:} \citet{FedAvg} propose the federated framework to corporately train a model among a cluster of edged devices with the isolated private dataset~\cite{root1}. To reduce the communication costs, multiple local updates and partial participation are yet adopted, which significantly undermines the performance~\cite{local_sgd} and becomes a trade-off in the federated framework in practice~\cite{avg_analysis, FedOpt,linear_speedup}. The essential reason for this is due to the global heterogeneous distribution of isolated private datasets~\cite {review3,review2, review,review4}. With further studies in FL, this is summarized as client drifts~\cite{SCAFFOLD} and inconsistency objectives among clients~\cite{inconsistency1,inconsistency2,inconsistency3, consistency}. To address these difficulties, a series of algorithms are proposed from the perspective of optimization. \citet{FedProx} propose to penalize the quadratic term of the equality constraint on the local objective to limit the local updates, which introduces a balance of local and global optima~\cite{L2GD}. \citet{FedSplit, FedDR} combine proximal mapping to approach the global consistency. \citet{VRLSGD, SCAFFOLD, FLcompress} utilize the variance reduction technique to reduce the stochastic variance of gradients caused by the sampling among clients. \citet{FedAdam} improve the performance by applying an efficient adaptive optimizer on the server side. \citet{FedPD} employ the primal-dual method on the server and clients respectively and \citet{FedADMM1, FedADMM2} extend the method to the Alternating Direction Method of Multipliers (ADMM). \citet{FedDyn} propose a variant of ADMM via dynamic regularization terms. \citet{FedSlowmo, FedGlomo} introduce the momentum to the local optimizer to improve the performance. \citet{FedCM, FedADC} bring the client-level momentum across the clients and global server which forces local updates to be similar.

\textbf{SAM:} \citet{Pre4SAM} firstly explore the strong relationship between the generalization and its flat loss landscape. Inspired by this, \citet{SAM} propose the SAM optimizer to search for a flat minimum with higher generality. \citet{ASAM} propose a scale-invariant SAM to improve the training stability. \citet{analysis4SAM} redefine the sharpness from an intuitive and theoretical view for SAM. After that, extensive variants of SAM are proposed \citet{penalize_gradient,sun2023adasam,zhong2022improving,mi2022make}. Existing works~\cite{SAM,ASAM} show that the flat loss landscape approached by the SAM optimizer holds higher stability and generality. Besides, \citet{FedSAM} firstly take into account the local generality to inherently enhance the consistency. \citet{improving_sam,sun2023fedspeed,shi2023make,shi2023improving} propose some variants to adopt the SAM optimizer in FL.
\section{Rethinking {\ttfamily FedSAM}  Algorithm}
\label{methodology}
Below, we introduce the preliminaries of FL and {\ttfamily FedSAM}. Then, we provide a deep understanding for {\ttfamily FedSAM}.

\subsection{Preliminaries}

\paragraph{Federated Learning (FL)}
Taking into account the classical FL framework, we focus on the minimization of the following finite sum problem:
\begin{equation}
\label{FL}
\begin{aligned}
    \mathop{\min}\limits_{w}{\Big\{} f(w) 
    &= \frac{1}{m}\sum_{i\in\left[\mathcal{m}\right]}f_i(w){\Big\}},\\
    f_i(w)
    &\triangleq \mathbb{E}_{\varepsilon_{i}\sim\mathcal{D}_{i}}f_{i}(w,\varepsilon_{i}),
\end{aligned}
\end{equation}
where $f\colon\mathbb{R}^{d}\rightarrow\mathbb{R}$ is the global objective function; $w$ is the parameters; $m$ is the number of total clients; $\left[\mathcal{m}\right]$ is the set of total clients; $\varepsilon_{i}$ is the data sample randomly drawn from the distribution $\mathcal{D}_i$ which may differ across the local clients as the heterogeneity; $f_i$ is the empirical loss of the $i$-th client.

\paragraph{Sharpness Aware Minimization (SAM)}
SAM aims to jointly minimize the loss function and smooth the loss landscape by solving the following problem:
\begin{equation}
\label{SAM}
    \mathop{\min}\limits_{w}{\Big\{}f_s(w)\triangleq\mathop{\max}\limits_{\Vert s\Vert\leq r}f(w+s){\Big\}},
\end{equation}
where $\Vert\cdot\Vert$ is $l_2$-norm and $r$ is the size of the neighborhood.\\
SAM does not solve the min-max objective and adopts an efficient approximation in practice. Via the first-order Taylor expansion of $f$, the solution of the inner maximization is:
\begin{equation}
    \begin{aligned}
        s^*(w)
        &\approx \mathop{\arg\max}\limits_{\Vert s\Vert\leq r} {\Big\{}f(w) + s^\top\nabla f(w){\Big\}}\\
        &= r\cdot \nabla f(w)/\Vert\nabla f(w)\Vert,
    \end{aligned}
\end{equation}
where $\nabla$ is the abbreviation for $\nabla_w$ on parameters $w$.\\ 
Substituting back into equation~(\ref{SAM}) and differentiating:
\begin{equation}
\label{SAM_g_approx}
    \begin{aligned}
        \nabla f_s(w)
        &\approx \nabla f(w+s^*(w))\\
        &\approx \nabla f(w)\vert_{w+r\cdot \nabla f(w)/\Vert\nabla f(w)\Vert}. 
    \end{aligned}
\end{equation}
Equation~(\ref{SAM_g_approx}) approximates the SAM gradient by the stage-wised calculation of 1) performing a gradient ascent step to reach the state $w+r\cdot \nabla f(w)/\Vert\nabla f(w)\Vert$ and calculating its gradient; 2) performing a gradient descent step with this gradient at state $w$. General SAM applies a standard numerical optimizer such as the Stochastic Gradient Descent (SGD) to approach the SAM objective~\cite{SAM}.

\subsection{Rethinking {\ttfamily FedSAM} Algorithm}
{\ttfamily FedSAM}~\cite{FedSAM} focuses on improving \textit{local generality} to achieve efficient training based on SAM optimizer for each client. According to equation~(\ref{FL})~(\ref{SAM}), by applying the SAM objective to replace the local ERM objective:
\begin{equation}
\label{FedSAM}
\begin{aligned}
    \mathop{\min}\limits_{w} {\Big\{}f^{\mathcal{s}}(w) 
    &= \frac{1}{m}\sum_{i\in\left[\mathcal{m}\right]}f^{\mathcal{s}}_i(w){\Big\}},\\
    f^{\mathcal{s}}_i(w)
    &\triangleq \mathop{\max}\limits_{\Vert s_i\Vert\leq r}f_i(w+s_i),
\end{aligned}
\end{equation}
where $s_i$ is the local perturbation allocated to $f_i$.\\
Its core idea is similar to personalized Federated Learning (pFL)~\cite{pFL1,pFL2}, which concerns more about improving the performance on the local dataset. Intuitively, the improvement of each client could promote better performance. Global Aggregation of local states with smoother local loss landscape helps boost the flatness of the global model. According to the formula~(\ref{SAM_g_approx}):
\begin{equation}
    \nabla f^{\mathcal{s}}(w)
    \approx \frac{1}{m}\sum_{i\in\left[\mathcal{m}\right]}\nabla f_i(w)\vert_{w+r\cdot \nabla f_i(w)/\Vert\nabla f_i(w)\Vert}. 
\end{equation}
The calculation of the global gradient can be easily split to each local client $i$, which allows {\ttfamily FedSAM} to directly enable the periodic aggregation from local training, like {\ttfamily FedAvg}.
\begin{figure}[b]
\centering
\vskip -0.1in
    \subfigure[High consistency.]{
        \includegraphics[width=0.24\textwidth]{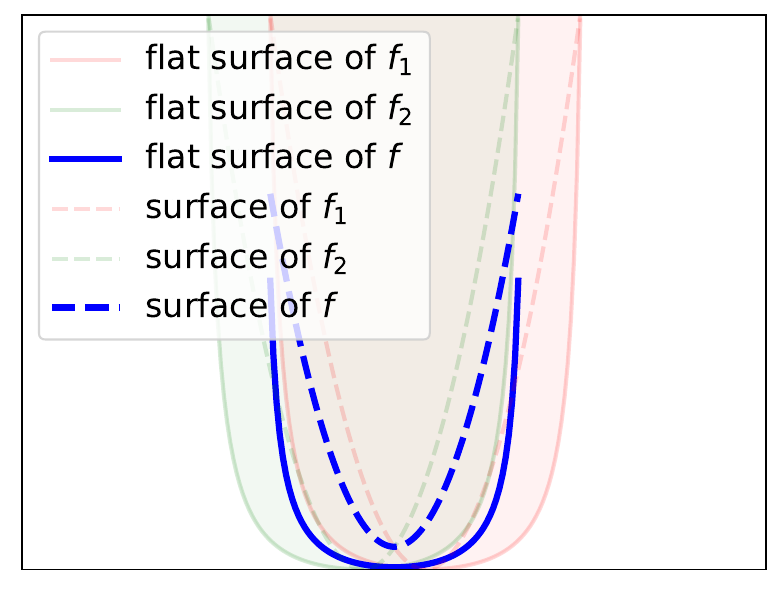}}\!\!\!\!
    \subfigure[Low consistency.]{
	\includegraphics[width=0.24\textwidth]{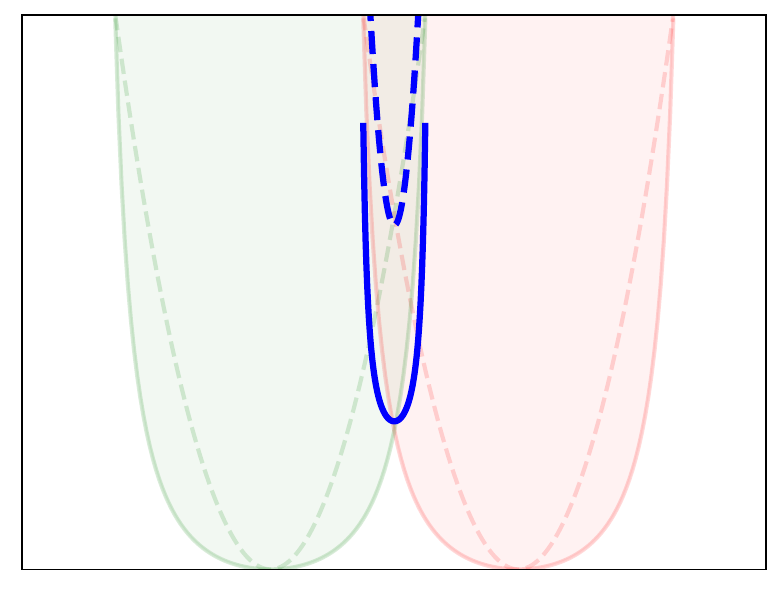}}
\vskip -0.05in
\caption{A toy schematic to introduce a bad case of {\ttfamily FedSAM}. We assume $m=2$ and $f=(f_1+f_2)/2$. The dotted lines represent the general loss surface trained by SGD, and the solid lines represent the flat loss surface trained by SAM. The red, green, and blue correspond to the local $f_1$, $f_2$, and global $f$, respectively.}
\label{toy}
\vskip -0.025in
\end{figure}
However, this indirect promotion also has its own limitations. Since the objective of equation~(\ref{FedSAM}) is not for the global function $f$, the flatness of the global landscape can not be directly optimized, which makes the whole process full of uncertainty. We illustrate a situation with the simple schematic shown in Figure~\ref{toy}. Obviously, when the local consistency maintains a high level, which means the local objectives are close to the global target, {\ttfamily FedSAM} effectively improves the smoothness of the global loss landscape like Figure~\ref{toy}~(a). However, when the consistency drops as shown in Figure~\ref{toy}~(b), though it contributes to reducing the loss value to a certain extent, the global loss surface is still sharp. An important reason is that the effective range of SAM usually cannot cover the total surface, and its performance drops significantly far away from the optima. Therefore, simply considering improving the local generality still faces very large limitations, e.g. rugged inconsistent local optimal under the severely heterogeneous dataset in practice.

\section{Methodology} 

To tackle the difficulties and improve the performance of the global function, we propose the {\ttfamily FedSMOO} which jointly considers both consistency and a global flat landscape. Then, we give convergence and generalization analysis.

\subsection{{\ttfamily FedSMOO} Algorithm}
In contrast to {\ttfamily FedSAM}~\cite{FedSAM},  we focus on adopting the SAM objective on the global function as follows:
\begin{equation}
\label{globalSAM}
\begin{aligned}
    \mathop{\min}\limits_{w} {\Big\{}\mathcal{F}(w) 
    &= \frac{1}{m}\sum_{i\in\left[\mathcal{m}\right]}\mathcal{F}_i(w){\Big\}},\\
    \mathcal{F}_i(w)
    &\triangleq \mathop{\max}\limits_{\Vert s\Vert\leq r}f_i(w+s),
\end{aligned}
\end{equation}
where $s$ is the global perturbation allocated to $f$.\\
In FL, due to the data privatization and localization, gradient calculations can only cover the local dataset of each client $i$. If directly adopting the equation~(\ref{globalSAM}), it requires the global gradient $\nabla f$ as mentioned in (\ref{SAM_g_approx}). This is impossible for the local clients. Therefore, we transform the objective~(\ref{globalSAM}) into the following form by adding two constraints on $w$ and $s$:
\begin{equation}
\label{our}
\begin{aligned}
    \mathop{\min}\limits_{w_i=w} {\Big\{}\mathcal{F} 
    &= \frac{1}{m}\sum_{i\in\left[\mathcal{m}\right]}\mathcal{F}_i(w,w_i,s,s_i){\Big\}},\\
    \mathcal{F}_i(w,w_i,s,s_i)
    &\triangleq \mathop{\max}_{
    \Vert s_i\Vert\leq r,
    \atop s_i = s}f_i(w_i+s_i).
\end{aligned}
\end{equation}
Under the constraints of $w_i$ and $s_i$, the objective~(\ref{our}) and (\ref{globalSAM}) are equivalent. Meanwhile, the parameters $w_i$ and perturbation $s_i$ could be easily split to each local client, which supports the federated calculations. This target guarantees the global consensus of $w_i$ and $s_i$, and could search for the consistent flat minima after completing the optimization.

\subsubsection{Re-approximating the Inner Maximization}
Like the general SAM, instead of solving this maximization problem rigorously, we derive an effective approximation of its solution. By taking the first-order Taylor expansion of $f_i(w_i+s_i)$, we define the local augmented Lagrangian function via penalizing the constrain $s_i=s$ as the linear and quadratic regularization terms on the objective as follows:
\begin{equation}
\label{AL4SAM}
\mathcal{L}_i^s : f_i(w_i)+s_i^\top\nabla f_i(w_i)+\mu_i^\top(s-s_i)+\frac{1}{2\alpha}\Vert s-s_i\Vert^2,
\end{equation}
where the local perturbation satisfies $\Vert s_i\Vert\leq r$.\\
To solve the maximization, we adopt a similar pattern as ADMM to alternately update the local perturbation $s_i$, the dual variable $\mu_i$, and global perturbation $s$. Firstly, we have:
\begin{equation}
\label{update_s_i}
\begin{aligned}
    \hat{s}_{i}
    &= \mathop{\arg\max}\limits_{\Vert s_i\Vert\leq r}{\Big\{}s_i^{\top}\left(\nabla f_{i}(w_i)\!-\!\mu_{i}\right)\!+\!\frac{1}{2\alpha}\Vert s_i\!-\!s\Vert^{2}{\Big\}}\\
    &= \mathop{\arg\max}\limits_{\Vert s_i\Vert\leq r}{\Big\{}\frac{1}{2\alpha}\Vert s_i + \mathcal{s}_i\Vert^{2}{\Big\}},
\end{aligned}
\end{equation}
where $\mathcal{s}_i=\alpha\left(\nabla f_{i}(w_i)-\mu_{i}\right)-s$. Thereby, $\hat{s}_{i}=r\mathcal{s}_i\ /\ \Vert \mathcal{s}_i\Vert$. Then we update the dual variable as $\mu_i=\mu_i+\frac{1}{\alpha}(\hat{s}_{i}-s)$.To maximize the global perturbation $s$, we have the following:
\begin{equation}
\label{update_s}
\begin{aligned}
    \hat{s}
    &= \mathop{\arg\max}\limits_{\Vert s\Vert\leq r}\frac{1}{m}\sum_{i\in\left[\mathcal{m}\right]}{\Big\{}s^{\top}{\mu_{i}}+\frac{1}{2\alpha}\Vert s-\hat{s}_{i}\Vert^{2}{\Big\}}\\
    &= \mathop{\arg\max}\limits_{\Vert s\Vert\leq r}{\Big\{}\frac{1}{2\alpha}\frac{1}{m}\sum_{i\in\left[\mathcal{m}\right]}\Vert s+\alpha\mu_i-\hat{s}_{i}\Vert^{2}{\Big\}}.
\end{aligned}
\end{equation}
Let $\mathcal{s}=\frac{1}{m}\sum_{i\in\left[\mathcal{m}\right]}\left(\alpha\mu_i-\hat{s}_{i}\right)$. Thereby, $\hat{s}=r\mathcal{s}\ /\ \Vert\mathcal{s}\Vert$.\\
With these update rules, we can calculate the local perturbation of approximately satisfying the constraints. Due to the data privatization and localization mentioned above, we can not directly get the $s^\star$ during the local training as its requirement of $s_i^\star$ for $i\in\left[\mathcal{m}\right]$. Therefore, we iterate the global perturbation on the global server and communicate it to the active clients at the new round. At the local training process, compared to the vanilla {\ttfamily FedSAM}, our proposed {\ttfamily FedSMOO} adopts the new perturbation vector corrected by the global estimation, which gradually approaches the global perturbation during the training. Then, according to the formula~(\ref{SAM_g_approx}), we get the new corrected gradient to perform optimization.

\subsubsection{Global Consistency and the Minimization}
As shown in Figure~\ref{toy}, to productively avoid the performance dropping and further improve the consistency, we also adopt a dynamic regularizer~\cite{FedDyn} on each local client via a similar pattern as ADMM, to efficiently minimize the global objective $\mathcal{F}$. Also, we penalize the $w_i=w$ constrain and introduce the global augmented Lagrangian function as:
\begin{equation}
\label{AL4Min}
\mathcal{L} : \frac{1}{m}\sum_{i}{\Big\{}\mathcal{F}_i+\lambda_i^{\top}(w^t-w_i)+\frac{1}{2\beta}\Vert w^t-w_i\Vert^{2}{\Big\}}.
\end{equation}
We split the finite sum problem to each local client and solve each sub-problem in one communication round for several local updates. In each sub-problem, we firstly minimize the local parameters $w_i$ in the augmented Lagrangian function,
\begin{equation}
\label{localupdate}
    w_{i,K}^t=\mathop{\arg\min}\limits_{w_i}{\Big\{}\mathcal{F}_i-\lambda_i^{\top}w_i+\frac{1}{2\beta}\Vert w^t-w_i\Vert^{2}{\Big\}}.
\end{equation}
In order not to affect the performance of the vanilla SAM, we adopt the SGD to solve this problem. Then we update the dual variable as $\lambda_i=\lambda_i - \frac{1}{\beta}(w_{i,K}^t-w^t)$. At the end of each round, we minimize the global variable $w^t$ from the function~(\ref{AL4Min}) and set it as $w^{t+1}$ to start the next round.

\subsubsection{Overview of {\ttfamily FedSMOO} Algorithm}
\begin{algorithm}[t]
	\renewcommand{\algorithmicrequire}{\textbf{Input:}}
	\renewcommand{\algorithmicensure}{\textbf{Output:}}
	\caption{{\ttfamily FedSMOO} Algorithm}
	\begin{algorithmic}[1]
		\REQUIRE global model $w$, local model $w_i$, communication round $T$, local interval $K$, local perturbation $s_i$, global perturbation $s$, dual variable for parameters $\lambda_i$, global dual variable $\lambda$, dual variable for perturbation $\mu_i$, penalized coefficient for the quadratic term $\alpha$, $\beta$.
		\ENSURE global model $w^T$.\\
            \STATE \textbf{Initialization} : $w=w^0$, $\lambda_i=\lambda=0$, $\mu_i=0$.
            \FOR{$t = 0, 1, 2, \cdots, T-1$}
            \STATE randomly select the active clients set $\left[\mathcal{n}\right]$ from $\left[\mathcal{m}\right]$
            \FOR{client $i \in \left[\mathcal{n}\right]$ \textbf{in parallel}}
            \STATE send the $w^t, s^{t}$ to the active clients as $w_{i,0}^{t}, s$
            \FOR{$ k=0, 1, \cdots, K-1$}
            \STATE get the stochastic gradient $g_{i,k}^{t}$ at $w_{i,k}^{t}$
            \STATE $\hat{s}_{i,k}^t=r\mathcal{s}_{i,k}^t\ /\ \Vert \mathcal{s}_{i,k}^t\Vert$ for $\mathcal{s}_{i,k}^t = (g_{i,k}^{t}\!-\mu_i)\!-\!s$
            \STATE $\mu_i = \mu_i + (\hat{s}_{i,k}^{t} - s)$
            \STATE get the stochastic gradient $\hat{g}_{i,k}^{t}$ at $w_{i,k}^{t}+\hat{s}_{i,k}^{t}$
            \STATE $w_{i,k+1}^t=w_{i,k}^{t}-\eta\big[\hat{g}_{i,k}^{t}-\lambda_i+\frac{1}{\beta}(w_{i,k}^{t}-w^t)\big]$
            \ENDFOR
            \STATE $\widetilde{s}_i = \mu_i-\hat{s}_{i,K}^t$
            \STATE $\lambda_i = \lambda_i - \frac{1}{\beta}(w_{i,K}^{t}-w^{t})$
            \STATE send the $w_{i}^{t}=w_{i,K}^{t}, \widetilde{s}_i$ to the global server
            \ENDFOR
            \STATE $s^t = r\mathcal{s}^t\ /\ \Vert \mathcal{s}^t\Vert$ for $\mathcal{s}^t = \frac{1}{n}\sum_{i\in \left[\mathcal{n}\right]}\widetilde{s}_i$
            \STATE $\lambda^{t+1} = \lambda^t - \frac{1}{\beta m}\sum_{i\in \left[\mathcal{n}\right]}(w_{i}^t-w^t)$
            \STATE $w^{t+1} = \frac{1}{n}\sum_{i\in \left[\mathcal{n}\right]}w_{i}^t-\beta \lambda^{t+1}$
            \ENDFOR
	\end{algorithmic}
	\label{algorithm}
\end{algorithm}

Algorithm~\ref{algorithm} shows our detailed implementation flow. At round $t$, we randomly select a sub-set of active clients $\left[\mathcal{n}\right]$ from the total clients set $\left[\mathcal{m}\right]$, and send the current global model $w^t$ and global perturbation estimation $s^t$ to all active clients. Line.8 computes the corrected local perturbation according to equation~(\ref{update_s_i}) with $\alpha=1$. Line.9 updates the dual variable for the local perturbation with the same $\alpha=1$. Line.10 computes the new corrected SAM gradient and Line.11 updates the local model $w_i$ with SGD according to equation~(\ref{localupdate}). Line.14 updates the local dual variable for the local model. After finishing the local training process, we send the required variables $w_i^t$ and $\widetilde{s}_i$ to the global server for aggregation. Line.17 computes the new global perturbation estimation according to the equation~(\ref{update_s}). Line.18 and Line.19 update the global model $w^{t+1}$ by minimizing the function~(\ref{AL4Min}) on the $w^t$. The global updates repeat $T$ rounds.

\subsection{Theoretical Analysis}
\label{theoretical analysis}
In this part, we demonstrate the theoretical analysis of our proposed {\ttfamily FedSMOO} Algorithm. We take into account the smooth non-convex case and prove its convergence rate can achieve fast $\mathcal{O}(1/T)$ with a general generalization bound.
\subsubsection{Convergence Rate}
Firstly we state some general assumptions in this work.
\begin{assumption}
\label{Lsmooth}
    Function $f_{i}(w)$ is $L$-smooth for all $i\!\in\!\left[\mathcal{m}\right]$, i.e., $\Vert\nabla f_{i}(x)-\nabla f_{i}(y)\Vert\leq L\Vert x-y\Vert$, for all $x,y\in\mathbb{R}^{d}$.
\end{assumption}
\begin{assumption}
\label{bounded_stochastic_gradient}
Stochastic gradient $g_{i,k}^{t}=\nabla f_{i}(w_{i,k}^{t}, \varepsilon_{i,k}^{t})$ with randomly sampled data $\varepsilon_{i,k}^{t}$ is an unbiased estimator of $\nabla f_{i}(w_{i,k}^t)$ with bounded variance, i.e., $\mathbb{E}[g_{i,k}^{t}]=\nabla f_{i}(w_{i,k}^{t})$ and $\mathbb{E}\Vert g_{i,k}^{t} - \nabla f_{i}(w_{i,k}^{t})\Vert^{2} \leq \sigma_{l}^{2}$, for all $\mathbf{x}_{i,k}^{t}\in\mathbb{R}^{d}$.
\end{assumption}
Assumption~\ref{Lsmooth} assumes the Lipschitz continuity of the gradients and Assumption~\ref{bounded_stochastic_gradient} assumes the bounded stochastic properties of the gradients. The two general assumptions are widely used in the analysis of the FL framework~\cite{FedAvg,FedAdam,SCAFFOLD,FedCM,FedDyn,FedSAM}. In addition to these, {\ttfamily FedSAM} requires a tighter bounded variance of the stochastic gradient and the assumption of bounded heterogeneity~\cite{FedSAM}. In our work, we refer to the approach of \citet{FedPD,FedDyn,FedADMM1,FedADMM2} and do not require the extra assumptions. Proof details can be referred to the $\textit{Appendix}$.

\begin{theorem}
\label{theorem1}
Let the assumptions hold, let the size of the active clients' set $\vert \left[\mathcal{n}\right]\vert=n$, and similarly, $\vert \left[\mathcal{m}\right]\vert=m$, let $r\leq \frac{4\kappa_r}{\sqrt{nT}}$ where $\kappa_r\in\mathbb{R}$ is a constant, and let $\beta\leq \frac{\sqrt{n}}{6\sqrt{6m}L}$, the sequence $\left\{\overline{w}^{t+1}\triangleq \frac{1}{n}\sum_{i\in\left[\mathcal{n}\right]}w_i^t\right\}_{t\in[T-1]}$ generated by the Algorithm~\ref{algorithm} under the non-convex case satisfy:
\begin{equation}
\begin{aligned}
    &\quad\frac{1}{T}\sum_{t=1}^{T}\mathbb{E}\Vert\nabla f(\overline{w}^t)\Vert^2\\
    &\leq \frac{1}{\zeta\beta T}\left[\kappa_f+\frac{1}{n}3\beta L\kappa_r+\frac{m}{n}72\beta^2L^2\delta^0\right],
\end{aligned}
\end{equation}
where $\zeta\in\left(0,\frac{1}{2}\right)$ is a constant, $\kappa_f\triangleq f(\overline{w}^1)-f^\star$ for $f^\star$ is the optima of $f$, $\delta^0\triangleq\frac{1}{m}\sum_{i\in\left[\mathcal{m}\right]}\mathbb{E}\Vert w_i^{0}-\overline{w}^{1}\Vert^{2}$ is the inconsistent term at the first round for $\overline{w}^{1}\triangleq \frac{1}{n}\sum_{i\in\left[\mathcal{n}\right]}w_i^0$.
\end{theorem}
\begin{remark}
Under the participation ratio equal to $n/m$, our proposed {\ttfamily FedSMOO} achieves fast $\mathcal{O}(1/T)$ convergence rate, which matches the conclusion of existing works~\cite{FedPD,FedDyn,FedADMM1,FedADMM2}. $\kappa_f$ term indicates the impact of the initial state $w^0$. $\kappa_r$ term comes from the corrected SAM steps. $\delta^0$ term is the variance of the local parameters $w_i$ optimized in the first round, which indicates the inconsistency level from the heterogeneity. The latter two achieve $n\times$ linear speedup.
\end{remark}
\begin{remark}
A key property is that local models will converge to the global model after training. With reference to the analysis of \citet{FedDyn,FedADMM1,FedADMM2} and consider the first-order gradient condition of equation~(\ref{localupdate}) as $\nabla \mathcal{F}_i(w_i^t)-\lambda_i+\frac{1}{\beta}(w_i^t-w^t)=0$. And, considering the update of $\lambda_i=\lambda_i-\frac{1}{\beta}(w_i^t-w^t)$, we infer $\lambda_i=\nabla \mathcal{F}_i(w_i^t)$ after updating at round $t$. Thus, rethinking the first-order gradient condition, we have the relationship of $\nabla \mathcal{F}_i(w_i^t)-\nabla \mathcal{F}_i(w_i^{t-1})+\frac{1}{\beta}(w_i^t-w^t)=0$. When $t\to \infty$, the local model $w_i^t\to w_i^\infty=w_i^{\infty-1}$ converges to the local state. Thus, we have $w_i^\infty\to w^\infty$. Furthermore, as shown in Theorem~\ref{theorem1}, $\overline{w}^t$ converges with the rate of $\mathcal{O}(1/t)$, which implies $\overline{w}^{\infty}=\frac{1}{n}\sum_{i\in\left[\mathcal{n}\right]}w_i^{\infty}\to w^{\infty}$. This matches the conclusion in~\cite{FedDyn} and guarantees that all local objectives converge to the consistent global objective.
\end{remark}

\subsubsection{Generalization Bound}
Based on the margin generalization bounds in~\cite{pac,generalization2}, we consider the generalization error bound on the global function $f$ as following:
\begin{equation}
\label{111}   G_\epsilon^f\triangleq\mathbf{P}\Big(f(w+s,\varepsilon)[y]\leq \max_{j\neq y}f(w+s,\varepsilon)[j]+\epsilon\Big),
\end{equation}
where $\varepsilon$ is the input data and $y$ is its ground truth. If $\epsilon=0$, we denote the equation~(\ref{111}) as $G^f$ to represent for the average misclassification rate under the global perturbation $s$. We define the $\widetilde{G}_\epsilon^f$ as the empirical estimate of the above expected margin loss trained on the heterogeneous dataset.

Different from the {\ttfamily FedSAM}~\cite{FedSAM}, which focuses on the \textit{averaged local generality} and define the error bound: $\frac{1}{m}\sum_i\mathbf{P}_i(f_i(w+s_i,\varepsilon)[y]\leq \mathop{\max}\limits_{j\neq y}f_i(w+s_i,\varepsilon)[j]+\epsilon)$. This split margin loss cannot describe the detailed characteristics of the global function $f$. To directly explore the stability of the global function, we adopt equation~(\ref{111}) as the margin  $\mathbf{P}(\frac{1}{m}\sum\limits_{i} f_i(w+s,\varepsilon)[y]\leq \mathop{\max}\limits_{j\neq y}\frac{1}{m}\sum\limits_{i}f_i(w+s,\varepsilon)[j]+\epsilon)$.

\begin{theorem}
\label{generalization_bound}
We assume the input data $\varepsilon$ as the normalized tensor, e.g. for an image, with the norm of $L_n$. And we consider a $L_l$-layer neural network with $d$ parameters per layer. The activation functions hold no biases and bounded 1-Lipschitz property. The input data will be bounded by a positive constant $L_w$ and its total size equal to $D$. Under the margin value of $\epsilon$ and a positive value $p$, with the probability of at least $1-p$, the empirical generalization risk on the global parameters $w$ generated by {\ttfamily FedSMOO} satisfies:
\begin{equation}
    G^f\leq \widetilde{G}_\epsilon^f + \mathcal{O}\left(\sqrt{\frac{L_l^2L_n^2d\ln(dL_l)\mathcal{V}_L+\ln\frac{L_lD}{p}}{(D-1)\epsilon^2}}\right),
\end{equation}
where $\mathcal{V}_L=\prod_{l=1}^{L}\Vert w_l\Vert^2\sum_{l=1}^L\frac{\Vert w_l\Vert_F^2}{\Vert w_l\Vert^2}$.
\end{theorem}
{\ttfamily FedSMOO} focuses on searching for the global consistent flat minima aligned to each local client, which theoretically guarantees the global consistency. Theorem~\ref{generalization_bound} indicates the upper bound of the generalization risk on the global parameters $w$ of the global function $f$ under SAM perturbation $s$, which guarantees the flat landscape of the global objective. Due to the space limitation, detailed proofs can be referred to the \textit{Appendix~C}.

\section{Experiments}
\label{experiments}

In this section, we firstly introduce our experimental implementation details, including the introduction of the baselines, the backbone, dataset splitting, and hyperparameters selection. Next, we state the evaluation of {\ttfamily FedSMOO} and study its hyperparameters' sensitivity and ablation test. 

\subsection{Implementation Details}
\label{setups}
\textbf{Baselines:} We compare the {\ttfamily FedSMOO} with several benchmarks: {\ttfamily FedAvg}~\cite{FedAvg} firstly introduce the federated framework via partial participation and local multiple training; {\ttfamily FedAdam}~\cite{FedAdam} implement the global adaptive optimizer; {\ttfamily SCAFFOLD}~\cite{SCAFFOLD} utilize the SVRG~\cite{SVRG} to diminish the client drift; {\ttfamily FedCM}~\cite{FedCM} maintain the consistency of local updates with a momentum term; {\ttfamily FedDyn}~\cite{FedDyn} guarantee the local collective objectives through the local dynamic regularization; {\ttfamily FedSAM} and its variant {\ttfamily MoFedSAM}~\cite{FedSAM} employ the local SAM~\cite{SAM} optimizer with momentum to reduce the generalized divergence drift. All methods are evaluated on ResNet-18~\cite{ResNet} implemented in PyTorch~\cite{PyTorch} official model-zoo. We follow the suggestion of \citet{replaceGN} to replace the Batch Normalization with the Group Normalization~\cite{GN} to avoid the non-differentiable parameters.

\textbf{Hyperparameter selections:} In the interests of the fairness, we freeze the general hyperparameters for all baselines on the same setups, including the local learning rate equal to $0.1$, the global learning rate equal to $1$ expect for {\ttfamily FedAdam} which adopts $0.1$ for the global adaptivity, the perturbation learning rate equal to $0.1$ expect for {\ttfamily FedSAM} and {\ttfamily MoFedSAM} which adopt $0.01$, the weight decay equal to 1$e$-3, and the learning rate decreasing by $0.998\times$ exponentially except for {\ttfamily FedDyn} and {\ttfamily FedSMOO} which adopt $0.9995\times$ for the proxy term. On the CIFAR-10, we take the batchsize equal to $50$, and the local epochs equal to $5$. On the CIFAR-100, we adjust the batchsize equal to $20$, and the local epochs equal to $2$ to avoid the extreme overfitting. More details of the selections can be found in \textit{Appendix~B}.

\textbf{Dataset splitting:} We conduct the experiments on CIFAR-$10/100$~\cite{CIFAR} and then follow the \citet{Dirichlet} to partition the dataset via widely used Dirichlet and Pathological sampling. The coefficient $u$ is adopted to control the variance, which measures the heterogeneity.

\begin{figure}[t]
\vskip 0.0in
\begin{center}
\centerline{\includegraphics[width=0.8\columnwidth]{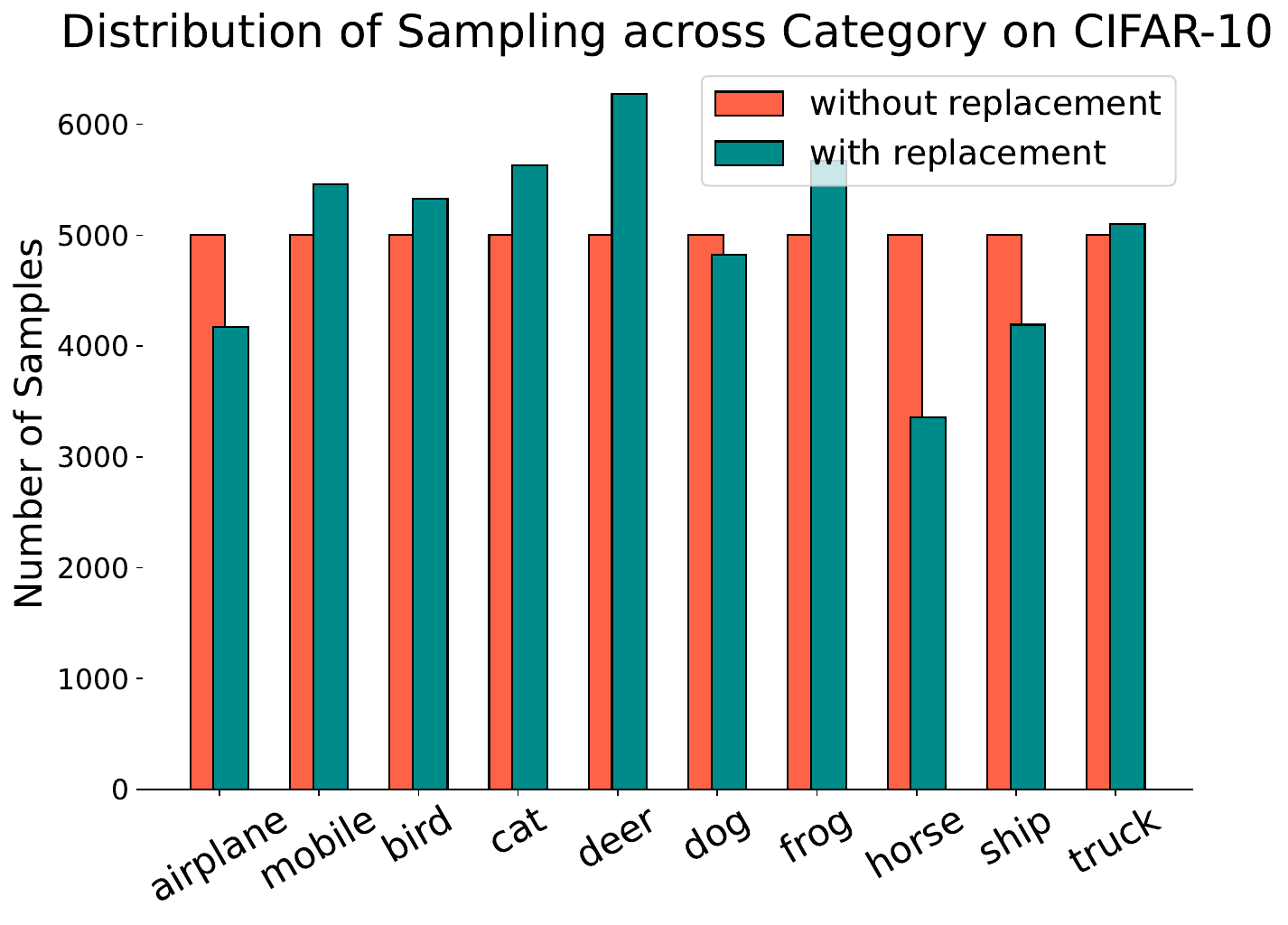}}
\vskip -0.05in
\caption{Distribution across category on CIFAR-10 of sampling with$/$without replacement under the Dirichlet coefficient $u=0.1$ and the number of total clients $m=100$. The standard deviation of the samples' number approaches 829.14, which highly increases their imbalance and properly approximates the practical scenes.}
\label{distribution_with_replacement}
\end{center}
\vskip -0.4in
\end{figure}

Additionally, to further enhance the heterogeneity and randomness, we enable the sampling \textit{with replacement} when generating the non-iid dataset. Most of the previous works only consider the local heterogeneity and sample the data \textit{without replacement}, which makes the distribution of the overall samples across the category uniform. In practice, due to the privacy protection policy, the distribution is usually disorderly and nonuniform. Some SOTA methods have significant performance degradation unanimously on this rugged dataset. Adopting sampling \textit{with replacement} breaks the balance of the raw data, which approaches more in line with real-world scenarios. Figure~\ref{distribution_with_replacement} demonstrates the difference between these two setups in the experiments.

\subsection{Experimental Evaluation}
\begin{table*}[t]
\begin{center}
\renewcommand{\arraystretch}{1}
\caption{Test accuracy comparison among baselines and our proposed method on the CIFAR-10/100 dataset after 800 rounds. The dataset splitting method is selected from the Dirichlet sampling with replacement and Pathological partition~(only a few random categories are enabled for sampling on a local client). The experimental setups are $10\%$-$100$ clients (upper part) and $5\%$-$200$ clients (lower part) respectively. $"u"$ represents the Dirichlet coefficient which is selected from $\left[ 0.1, 0.6 \right]$, and $"c"$ represents the number of active categories on each client which is selected from $\left[ 3, 6 \right]$ on CIFAR-10 and $\left[ 10, 20 \right]$ on CIFAR-100. Each result is calculated by $2$ times.}
\vspace{0.07cm}
\begin{sc}
\small
\setlength{\tabcolsep}{2.67mm}{\begin{tabular}{@{}c|cccccccc@{}}
\toprule
\multicolumn{1}{c}{\multirow{3}{*}{Algorithm}} & \multicolumn{4}{c}{CIFAR-10} & \multicolumn{4}{c}{CIFAR-100} \\
\cmidrule(lr){3-4} \cmidrule(lr){7-8}
\multicolumn{1}{c}{} & \multicolumn{2}{c}{Dirichlet} & \multicolumn{2}{c}{Pathological} & \multicolumn{2}{c}{Dirichlet} & \multicolumn{2}{c}{Pathological} \\ 
\cmidrule(lr){2-5} \cmidrule(lr){6-9}
\multicolumn{1}{c}{} & \multicolumn{1}{c}{$u=0.6$} & \multicolumn{1}{c}{$u=0.1$} & \multicolumn{1}{c}{$c=6$} & \multicolumn{1}{c}{$c=3$} & \multicolumn{1}{c}{$u=0.6$} & \multicolumn{1}{c}{$u=0.1$} & \multicolumn{1}{c}{$c=20$} & \multicolumn{1}{c}{$c=10$} \\
\cmidrule(lr){1-1} \cmidrule(lr){2-9}
{\ttfamily FedAvg}       & $\text{79.52}_{\pm.13}$ & $\text{76.00}_{\pm.18}$ & $\text{79.91}_{\pm.17}$ & $\text{74.08}_{\pm.22}$ & $\text{46.35}_{\pm.15}$ & $\text{42.64}_{\pm.22}$ & $\text{44.15}_{\pm.17}$ & $\text{40.23}_{\pm.31}$ \\ 
{\ttfamily FedAdam}      & $\text{77.08}_{\pm.31}$ & $\text{73.41}_{\pm.33}$ & $\text{77.05}_{\pm.26}$ & $\text{72.44}_{\pm.29}$ & $\text{48.35}_{\pm.17}$ & $\text{40.77}_{\pm.31}$ & $\text{41.26}_{\pm.30}$ & $\text{32.58}_{\pm.22}$ \\ 
{\ttfamily SCAFFOLD}     & $\text{81.81}_{\pm.17}$ & $\text{78.57}_{\pm.14}$ & $\text{83.07}_{\pm.10}$ & $\text{77.02}_{\pm.18}$ & $\text{51.98}_{\pm.23}$ & $\text{44.41}_{\pm.15}$ & $\text{46.06}_{\pm.22}$ & $\text{41.08}_{\pm.24}$ \\ 
{\ttfamily FedCM}        & $\text{82.97}_{\pm.21}$ & $\text{77.82}_{\pm.16}$ & $\text{83.44}_{\pm.17}$ & $\text{77.82}_{\pm.19}$ & $\text{51.56}_{\pm.20}$ & $\text{43.03}_{\pm.26}$ & $\text{44.94}_{\pm.14}$ & $\text{38.35}_{\pm.27}$ \\ 
{\ttfamily FedDyn}       & $\text{83.22}_{\pm.18}$ & $\text{78.08}_{\pm.19}$ & $\text{83.18}_{\pm.17}$ & $\text{77.63}_{\pm.14}$ & $\text{50.82}_{\pm.19}$ & $\text{42.50}_{\pm.28}$ & $\text{44.19}_{\pm.19}$ & $\text{38.68}_{\pm.14}$ \\
{\ttfamily FedSAM}       & $\text{80.10}_{\pm.12}$ & $\text{76.86}_{\pm.16}$ & $\text{80.80}_{\pm.23}$ & $\text{75.51}_{\pm.24}$ & $\text{47.51}_{\pm.26}$ & $\text{43.43}_{\pm.12}$ & $\text{45.46}_{\pm.29}$ & $\text{40.44}_{\pm.23}$ \\ 
{\ttfamily MoFedSAM}     & $\text{84.13}_{\pm.13}$ & $\text{78.71}_{\pm.15}$ & $\text{84.92}_{\pm.14}$ & $\text{79.57}_{\pm.18}$ & $\textbf{54.38}_{\pm.22}$ & $\text{44.85}_{\pm.25}$ & $\text{47.42}_{\pm.26}$ & $\text{41.17}_{\pm.22}$ \\ 
{\ttfamily \textbf{Our}} &$\textbf{84.55}_{\pm.14}$&$\textbf{80.82}_{\pm.17}$&$\textbf{85.39}_{\pm.21}$&$\textbf{81.58}_{\pm.16}$&$\text{{53.92}}_{\pm.18}$&$\textbf{46.48}_{\pm.13}$&$\textbf{48.87}_{\pm.17}$&$\textbf{44.10}_{\pm.19}$\\
\cmidrule(lr){1-1} \cmidrule(lr){2-5} \cmidrule(lr){6-9}
{\ttfamily FedAvg}       & $\text{75.90}_{\pm.21}$ & $\text{72.93}_{\pm.19}$ & $\text{77.47}_{\pm.34}$ & $\text{71.86}_{\pm.34}$ & $\text{44.70}_{\pm.22}$ & $\text{40.41}_{\pm.33}$ & $\text{38.22}_{\pm.25}$ & $\text{36.79}_{\pm.32}$ \\  
{\ttfamily FedAdam}      & $\text{75.55}_{\pm.38}$ & $\text{69.70}_{\pm.32}$ & $\text{75.24}_{\pm.22}$ & $\text{70.49}_{\pm.26}$ & $\text{44.33}_{\pm.26}$ & $\text{38.04}_{\pm.25}$ & $\text{35.14}_{\pm.16}$ & $\text{30.28}_{\pm.28}$ \\ 
{\ttfamily SCAFFOLD}     & $\text{79.00}_{\pm.26}$ & $\text{76.15}_{\pm.15}$ & $\text{80.69}_{\pm.21}$ & $\text{74.05}_{\pm.31}$ & $\text{50.70}_{\pm.18}$ & $\text{41.83}_{\pm.29}$ & $\text{39.63}_{\pm.31}$ & $\text{37.98}_{\pm.36}$ \\ 
{\ttfamily FedCM}        & $\text{80.52}_{\pm.29}$ & $\text{77.28}_{\pm.22}$ & $\text{81.76}_{\pm.24}$ & $\text{76.72}_{\pm.25}$ & $\text{50.93}_{\pm.31}$ & $\text{42.33}_{\pm.19}$ & $\text{42.01}_{\pm.17}$ & $\text{38.35}_{\pm.24}$ \\ 
{\ttfamily FedDyn}       & $\text{80.69}_{\pm.23}$ & $\text{76.82}_{\pm.17}$ & $\text{82.21}_{\pm.18}$ & $\text{74.93}_{\pm.22}$ & $\text{47.32}_{\pm.18}$ & $\text{41.74}_{\pm.21}$ & $\text{41.55}_{\pm.18}$ & $\text{38.09}_{\pm.27}$ \\
{\ttfamily FedSAM}       & $\text{76.32}_{\pm.16}$ & $\text{73.44}_{\pm.14}$ & $\text{78.16}_{\pm.27}$ & $\text{72.41}_{\pm.29}$ & $\text{45.98}_{\pm.27}$ & $\text{40.22}_{\pm.27}$ & $\text{38.71}_{\pm.23}$ & $\text{36.90}_{\pm.29}$ \\ 
{\ttfamily MoFedSAM}     & $\text{82.58}_{\pm.21}$ & $\text{78.43}_{\pm.24}$ & $\text{84.46}_{\pm.20}$ & $\text{79.93}_{\pm.19}$ & $\textbf{53.51}_{\pm.25}$ & $\text{42.22}_{\pm.23}$ & $\text{42.77}_{\pm.27}$ & $\text{39.81}_{\pm.21}$ \\ 
{\ttfamily \textbf{Our}} &$\textbf{82.94}_{\pm.19}$&$\textbf{79.76}_{\pm.19}$&$\textbf{84.82}_{\pm.18}$&$\textbf{81.01}_{\pm.19}$&$\text{{53.45}}_{\pm.19}$&$\textbf{45.83}_{\pm.18}$&$\textbf{44.70}_{\pm.21}$&$\textbf{43.41}_{\pm.22}$\\
\bottomrule
\end{tabular}}
\label{acc}
\end{sc}
\end{center}
\vspace{-0.4cm}
\end{table*}
\textbf{Comparison with the baselines:}
 As shown in Table~\ref{acc}, our proposed {\ttfamily FedSMOO} performs well with good stability and effectively resists the negative impact of the strong heterogeneous dataset. Specifically, on the CIFAR-10 dataset, {\ttfamily FedSMOO} achieves $80.82\%$ on the Dirichlet-$0.1$ setups, which is $2.11\%$ higher than the second highest accuracy. On the CIFAR-100 dataset, the improvement achieves $1.63\%$ from the {\ttfamily MoFedSAM}. Improvements of the vanilla {\ttfamily FedSAM} on both CIFAR10 / 100 are very limited, which is only about $1\%$ over than {\ttfamily FedAvg} on average. While its variant {\ttfamily MoFedSAM} achieves about $5\%$ improvement over the {\ttfamily FedSAM} on average, which benefits from the momentum updates during the local training. {\ttfamily FedSMOO} focuses on both the optimization and global generalization performance, and searches for the consistent flat landscape, which efficiently converges to a better minimum and is significantly ahead of the {\ttfamily MoFedSAM} algorithm, especially on the severely heterogeneous dataset with inconsistent local optima.

\textbf{Impact of heterogeneity:} 
Via sampling \textit{with replacement}, not only is there a data imbalance between local clients, but also the number of samples among categories in the global dataset is also very different, as shown in Figure~\ref{distribution_with_replacement}. With the larger heterogeneity, several baselines are greatly affected. We select the two methods to split the data, the Dirichlet and the Pathological. In the Dirichlet, we select the variance coefficient equal to $0.1$ and $0.6$ (the results of $0.6$ and iid are very close). In detail, on CIFAR-10, when the coefficient decreases from $0.6$ to $0.1$, {\ttfamily MoFedSAM} drops from $84.13\%$ to $78.71\%$, while {\ttfamily FedSMOO} drops about only $3.73\%$ to $80.82\%$, which maintains high accuracy. On CIFAR-100, {\ttfamily MoFedSAM} achieves a little higher accuracy on Dirichlet-$0.6$ low heterogeneous dataset, while it severely drops about $9.53\%$ when the non-iid level decreases to Dirichlet-$0.1$. {\ttfamily FedSMOO} works well on the high heterogeneity and achieves $1.63\%$ ahead of the second-highest accuracy. Furthermore, we test the Pathological split by blocking certain types on a client and only sampling from the total random $c$ categories, which is a higher level of heterogeneity. On the pathological dataset with only 3 categories per client, {\ttfamily FedSMOO} is $2.01\%$ ahead of the {\ttfamily MoFedSAM} on the $10\%$ participation and $1.08\%$ on $5\%$ participation. It achieves good performance on the strong heterogeneous dataset with high stability on different setups.

\begin{figure*}[t]
\centering
    \subfigure[{\ttfamily FedAvg} $v.s$ {\ttfamily FedSMOO}]{
        \includegraphics[width=0.3\textwidth]{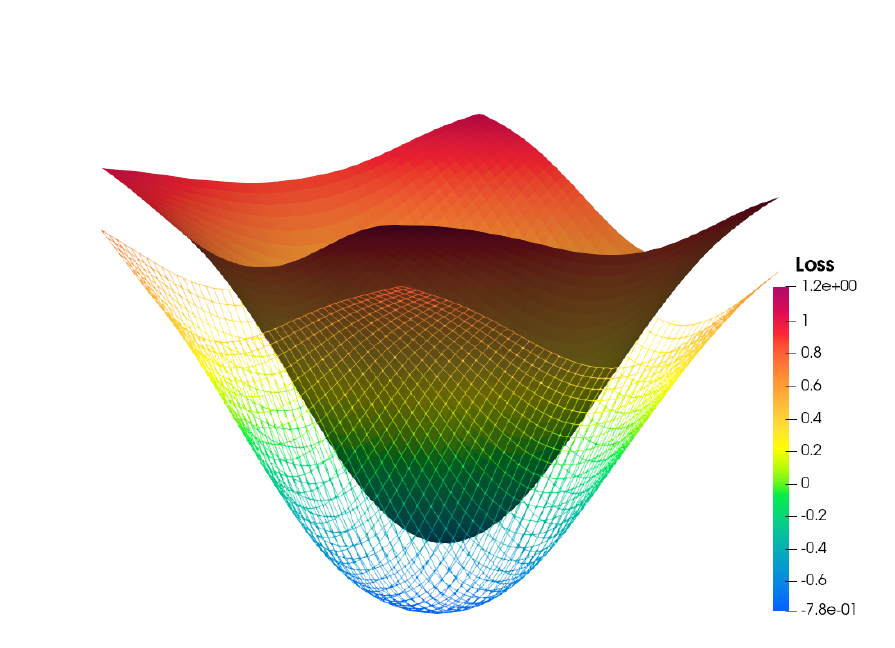}}\!\!
    \subfigure[{\ttfamily SCAFFOLD} $v.s$ {\ttfamily FedSMOO}]{
	\includegraphics[width=0.3\textwidth]{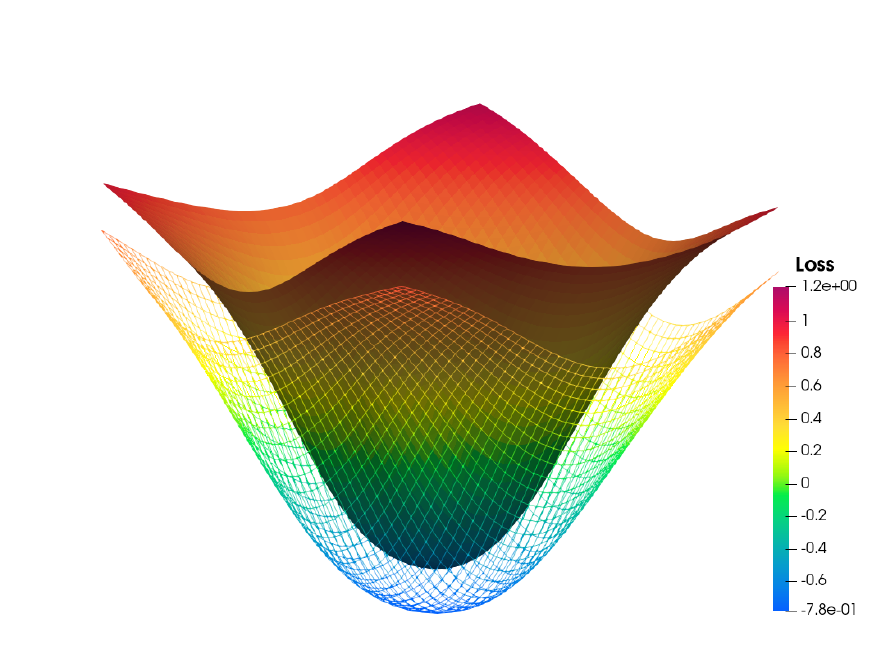}}\!\!
    \subfigure[{\ttfamily MoFedSAM} $v.s$ {\ttfamily FedSMOO}]{
	\includegraphics[width=0.3\textwidth]{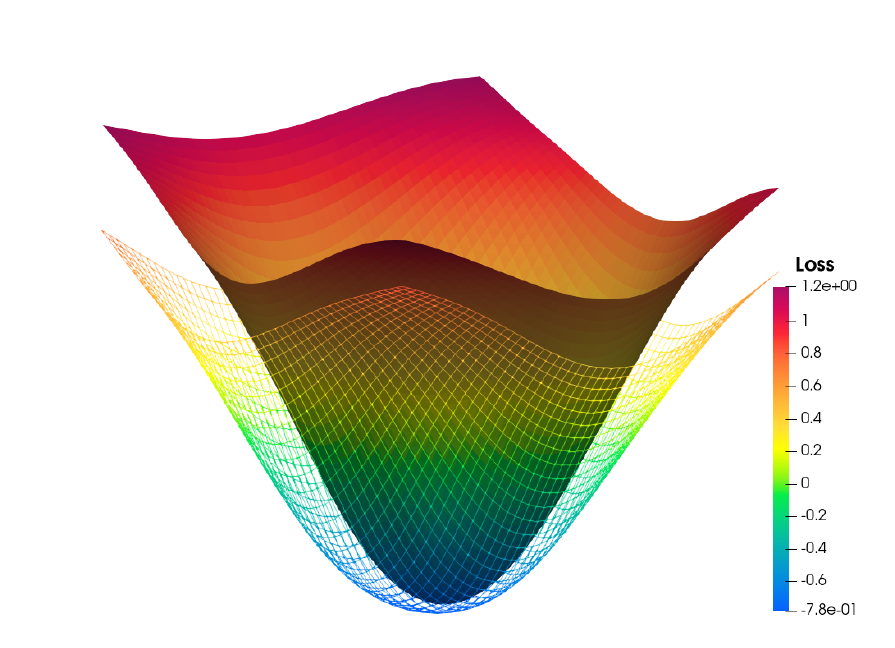}}
\vskip -0.05in
\caption{Visualization of the loss landscape of ResNet-18 backbone trained via {\ttfamily FedAvg}, {\ttfamily SCAFFOLD}, {\ttfamily MoFedSAM} and {\ttfamily FedSMOO} on the CIFAR-10 dataset. For clarity, we use the grid surface on the {\ttfamily FedSMOO} and compare it with the other three baselines separately. {\ttfamily FedSMOO} could approach a more general and flat loss landscape which efficiently improves the generalization performance in FL.}
\label{surface-comparison}
\end{figure*}
\begin{figure*}[t]
\vskip -0.05in
\centering
    \subfigure[Different local interval.]{
        \includegraphics[width=0.24\textwidth]{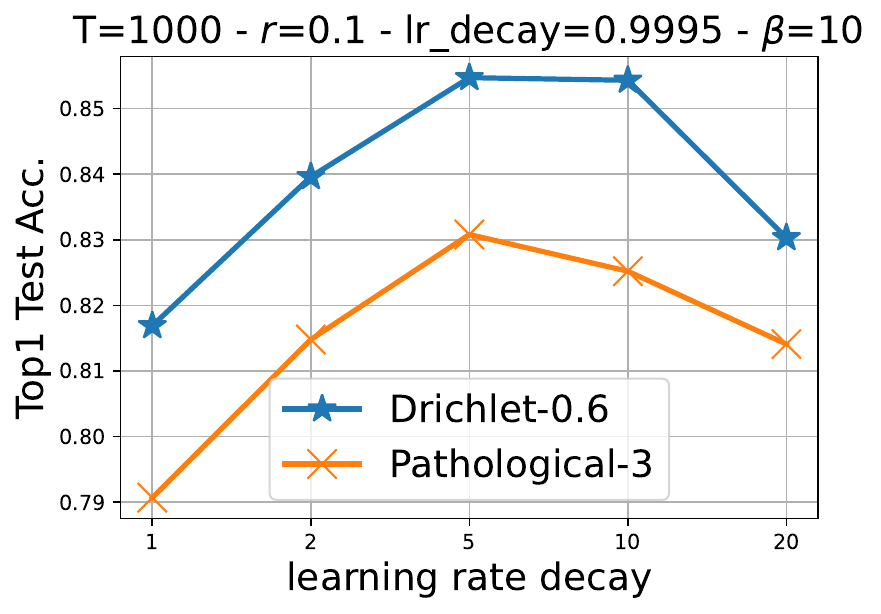}}
    \subfigure[Selection of lr-decay.]{
	\includegraphics[width=0.24\textwidth]{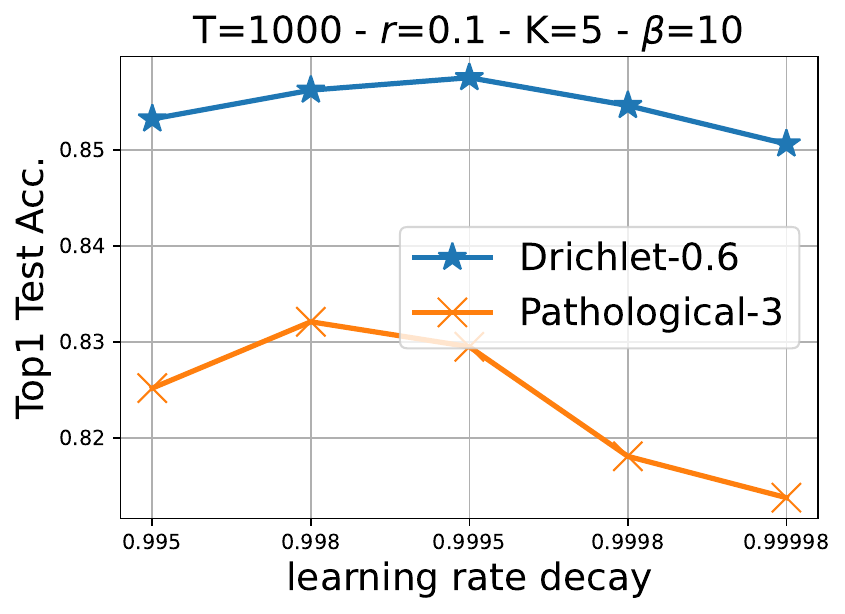}}
    \subfigure[Selection of $\beta$.]{
	\includegraphics[width=0.24\textwidth]{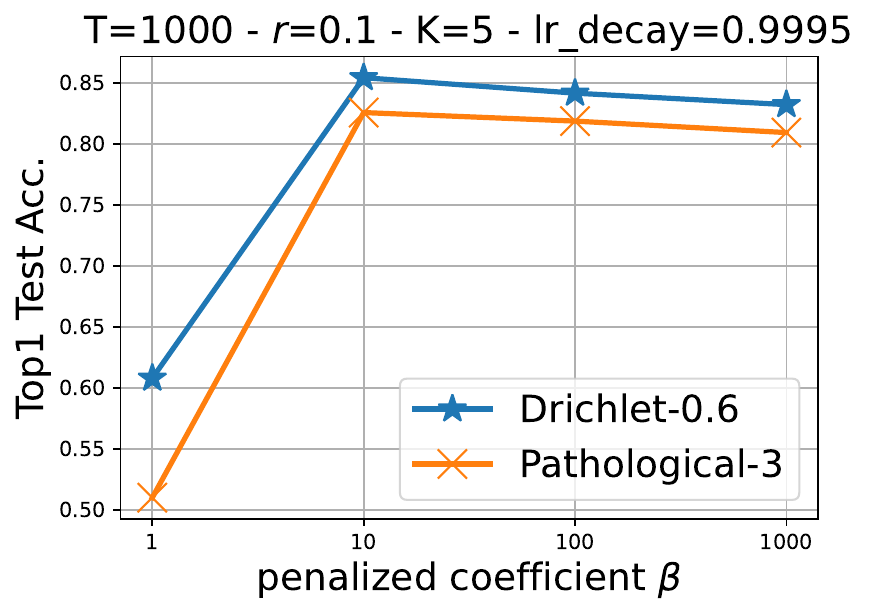}}
    \subfigure[Selection of $r$.]{
	\includegraphics[width=0.24\textwidth]{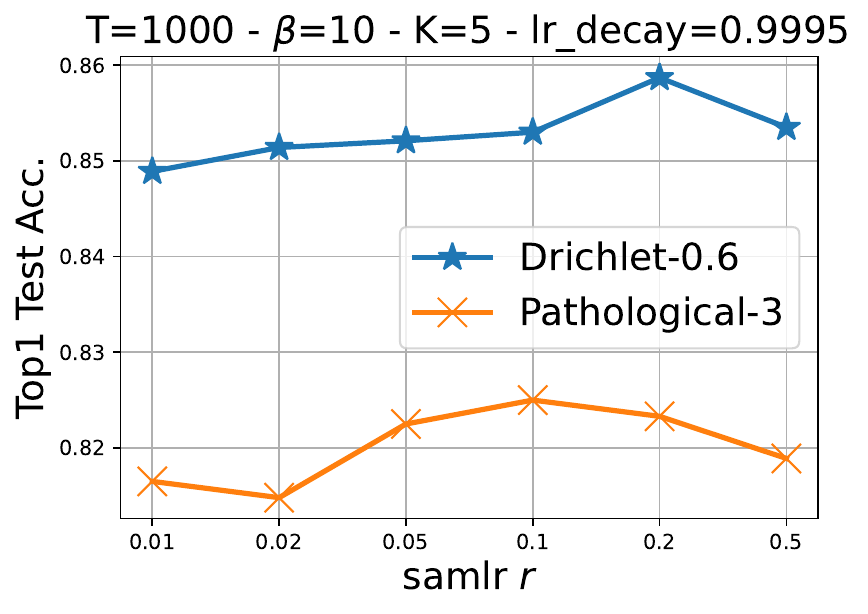}}
\vskip -0.1in
\caption{Hyperparameters sensitivity studies of local intervals, learning rate decay, penalized coefficient $\beta$ and SAM-lr $r$ on CIFAR-10.}
\label{hyperparamters}
\vskip -0.2in
\end{figure*}

\textbf{Impact of partial participation:}
Another important option is the partial participation ratio. To fairly compare with the baselines, we froze the other hyperparameters selections. When the participation ratio decreases from $10\%$ to $5\%$ on CIFAR-10, {\ttfamily FedSMOO} still achieves excellent performance, which drops from $80.82\%$ to $79.76\%$ on Dirichlet-$0.1$, and from $81.58\%$ to $81.01\%$ on Pathological-$3$. On the CIFAR-100 experiments, {\ttfamily FedSMOO} maintains a high level of generalization on Pathological-$10$, which achieves $44.10\%$.

\textbf{Loss landscape:}
Figure~\ref{surface-comparison} shows the visualization of the loss landscape. All models are trained on CIFAR-10 with the Dirichlet-$0.1$ setup. We compare the 
{\ttfamily FedAvg}, {\ttfamily SCAFFOLD}, {\ttfamily MoFedSAM} and {\ttfamily FedSMOO} algorithms. For a clear comparison, we superimpose the landscape of the two algorithms together and apply the grid surface on {\ttfamily FedSMOO}. It can be seen that the global minima approached by our method has not only a lower loss value but also a smoother loss landscape, than the other three benchmarks.

\textbf{Hyperparameters Sensitivity:}
We study the hyperparameters' sensitivity of local interval, learning rate decay, penalized coefficient $\beta$, and SAM learning rate $r$. {\ttfamily FedSMOO} is stable on the changes in learning rate decay and $r$. The selection of local intervals needs to be long enough to approximate the solution of Equation~(\ref{localupdate}) as mentioned in \citet{FedDyn}.
The selection can be fine-tuned appropriately for the gain. The penalized coefficient $\beta$ can not be selected too small, which will cause the proxy term to severely interfere with the direction of gradient descent.

\begin{table}[b]
\small
    \vspace{-0.4cm}
    \caption{Consistency and Hessian matrix.}
    \label{hessian}
    \vspace{0.1cm}
    \centering
    \setlength{\tabcolsep}{1.5mm}{\begin{tabular}{c|cccc}
    \toprule
     method & \multicolumn{2}{c}{{\ttfamily FedSAM}} & \multicolumn{2}{c}{{\ttfamily FedSMOO}} \\
     \midrule
     Dirichlet & 0.6 & 0.1 & 0.6 & 0.1 \\
     \midrule
     $\frac{1}{n}\sum_i\Vert w_i^t-w^t\Vert^2$ & 0.866 & 1.245 & 0.821 & 1.061 \\
     Hessian Top Eigenvalue & 142.65 & 177.18 & 91.46 & 107.44  \\
     Hessian Trace & 3104.1 & 3842.3 & 1783.3 & 2689.4 \\
    \bottomrule
\end{tabular}}
\end{table}
\textbf{Hessian:}
To further illustrate the efficiency of improving the generalization, we test the Hessian matrix of the final models. We test ResNet-18-GN on the CIFAR-10 dataset with the setup of 10\%-100 clients under the Dirichlet sampling coefficient equal to 0.1 and 0.6~(the same hyperparameter selection in Table~\ref{acc}. The results are shown in Table~\ref{hessian}. It is clear to see that the proposed {\ttfamily FedSMOO} approaches the flat minimal with a lower Hessian top eigenvalue than the {\ttfamily FedSAM} method. The Hessian trace also decreases effectively. The divergence term $\frac{1}{n}\sum_i\Vert w_i^t-w^{t+1}\Vert^2$ maintains similar properties with the flatness in our experiments.

\textbf{Communication Cost:}
One of the major concerns in FL is the communication bottleneck. Though some advanced algorithms achieve higher performance, they have to communicate more data to support the training process. This is undoubtedly a very large limitation in the FL paradigm. Therefore, we compare the communication costs for the algorithms in our paper. For convenience, we assume {\ttfamily FedAvg} communicates total $V$ bits in the training process. To achieve 74\% on CIFAR-10 of Pathological-3 splitting, the total communication costs are stated in Table~\ref{communication}.
\begin{table}[H]
\small
    \vspace{-0.2cm}
    \caption{Total communication costs.}
    \label{communication}
    \vspace{0.1cm}
    \centering
    \setlength{\tabcolsep}{1.5mm}{\begin{tabular}{c|cc}
    \toprule
     method & rounds & communication costs \\
     \midrule
     {\ttfamily FedAvg}   & 723~(1$\times$) & $V$ \\
     {\ttfamily FedAdam}  & - & - \\
     {\ttfamily SCAFFOLD} & 533~(1.36$\times$) & 1.47 $V$ \\
     {\ttfamily FedCM}    & 442~(1.64$\times$) & 1.22 $V$ \\
     {\ttfamily FedDyn}   & 424~(1.71$\times$) & 0.59 $V$ \\
     {\ttfamily FedSAM}   & 604~(1.20$\times$) & 0.84 $V$ \\
     {\ttfamily MoFedSAM} & 298~(2.43$\times$) & 0.82 $V$ \\
     {\ttfamily FedSMOO}  & 194~(3.73$\times$) & 0.53 $V$ \\
    \bottomrule
\end{tabular}}
\vspace{-0.2cm}
\end{table}

From this table, we clearly see that advanced methods, i.e. {\ttfamily SCAFFOLD} and {\ttfamily FedCM}, suffer from very high communication costs for transferring double vectors per round. Though our proposed {\ttfamily FedSMOO} requires double as well, it focuses on improving the generalization performance in FL and effectively reduces the total communication costs.


\section{Conclusion}
\label{conclusion}
In this work, we propose a novel and practical federated algorithm {\ttfamily FedSMOO} which jointly considers the optimization and generalization targets via adopting the dynamic regularizer to guarantee the local optima towards the global objective revised by the global SAM optimizer. It efficiently searches for a consistent flat minimum in the FL framework. Theoretical analysis guarantees that {\ttfamily FedSMOO} achieves the fast convergence rate of $\mathcal{O}(1/T)$. Furthermore, we also provide the global generalization bound. We conduct extensive experiments to verify its efficiency and significant performance on the global model under severe heterogeneity.

\paragraph{Acknowledgement.} {\small Prof. Dacheng Tao is partially supported by Australian Research Council Project FL-170100117.
}

\bibliography{example_paper}
\bibliographystyle{icml2023}

\clearpage
\appendix
\onecolumn
\section{Appendix}
In this part, we provide the appendix supplementary materials including some experimental results and the proof of the main theorem.

\section{Experiments}
\label{appendix:experiments}
In this part, we present the results of some complementary experiments.

\subsection{Introduction of the Dataset}
\begin{table}[h]
  \caption{Dataset introductions.}
  \vskip 0.1in
  \label{dataset}
  \centering
  \begin{tabular}{ccccc}
    \toprule
    Dataset     & Training Data     & Test Data & Class & Size\\
    \midrule
    CIFAR-10 & 50,000  & 10,000 & 10 & 3$\times$32$\times$32    \\
    CIFAR-100     & 50,000 & 10,000 & 100 & 3$\times$32$\times$32    \\
    \bottomrule
  \end{tabular}
\end{table}
CIFAR-10 / 100 are two basic dataset in the computer version studies. Data samples in both are the colorful images with the small resolution of $32\times32$. The main reason why FL pays more attention to this kind of small dataset is that FL focuses on privacy-protection training on isolated small dataset, e.g. for medical images. Usually the resolution of such data is not high, and the number of samples per class is very limited due to its expensive labeling cost.

\subsection{Distributions of Dirichlet and Pathological Split}
\begin{figure}[ht]
\vskip -0.2in
\centering
    \subfigure[Dirichlet-0.6 on CIFAR-100.]{
        \includegraphics[width=0.48\textwidth]{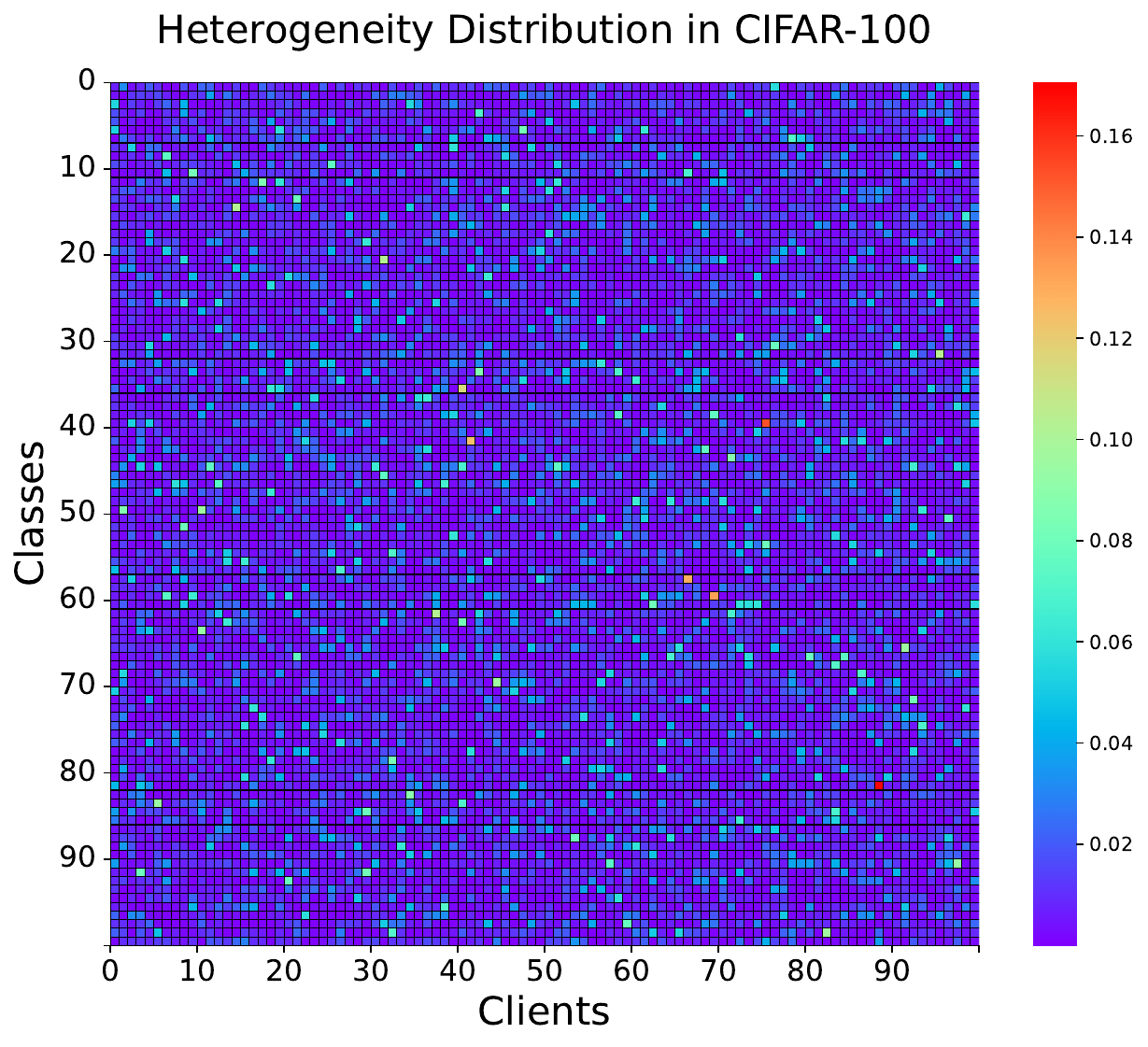}}
    \subfigure[Pathological-30 on CIFAR-100.]{
	\includegraphics[width=0.48\textwidth]{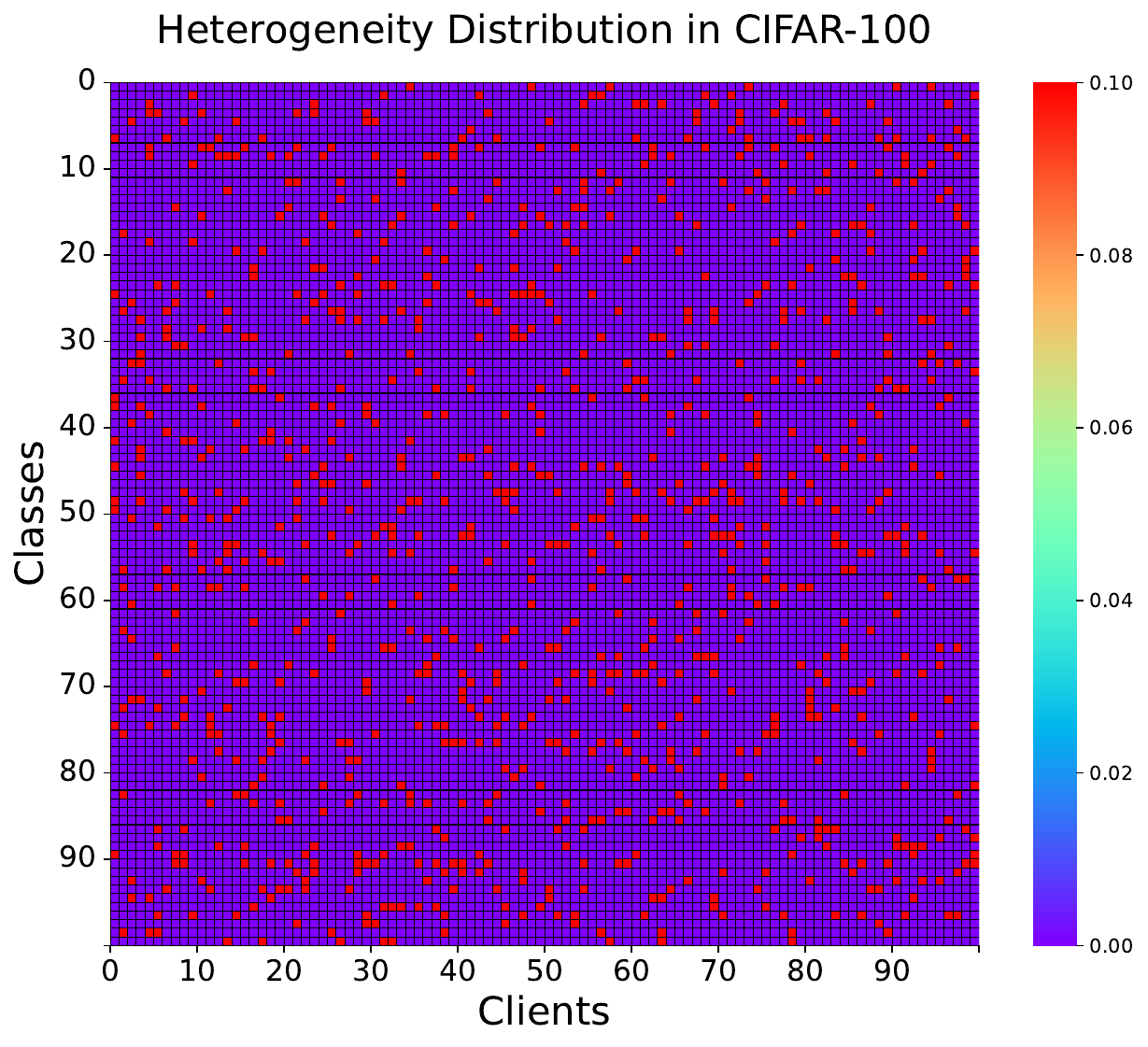}}
\vskip -0.05in
\caption{Heat-map of the Dirichlet split and Pathological split.}
\label{distributions_of_different_split}
\vskip -0.05in
\end{figure}
Above figures show the different distribution of the Dirichlet-0.6 and Pathological-30 on CIFAR-100 dataset. The main differences are:
\begin{itemize}
    \item \textbf{Dirichlet:} Each category can be sampled with a non-zero probability. The local dataset obeys a Dirichlet distribution. As shown in Figure~\ref{distributions_of_different_split}~(a), each category has a color.
    \item \textbf{Pathological:} Only selected categories can be sampled with a non-zero probability. The local dataset obeys a uniform distribution of active categories. As shown in Figure~\ref{distributions_of_different_split}~(b), the category only has a color of purple or red.
\end{itemize}

\subsection{Sampling \textit{with replacement}}
We have shown the difference between the difference between the vanilla process and our process. Adopting sampling \textit{with replacement} will extremely change the distribution of the total dataset. Here we also show their differences in CIFAR-100 dataset.\\
\begin{figure}[ht]
\vskip 0.0in
\begin{center}
\centerline{\includegraphics[width=\columnwidth]{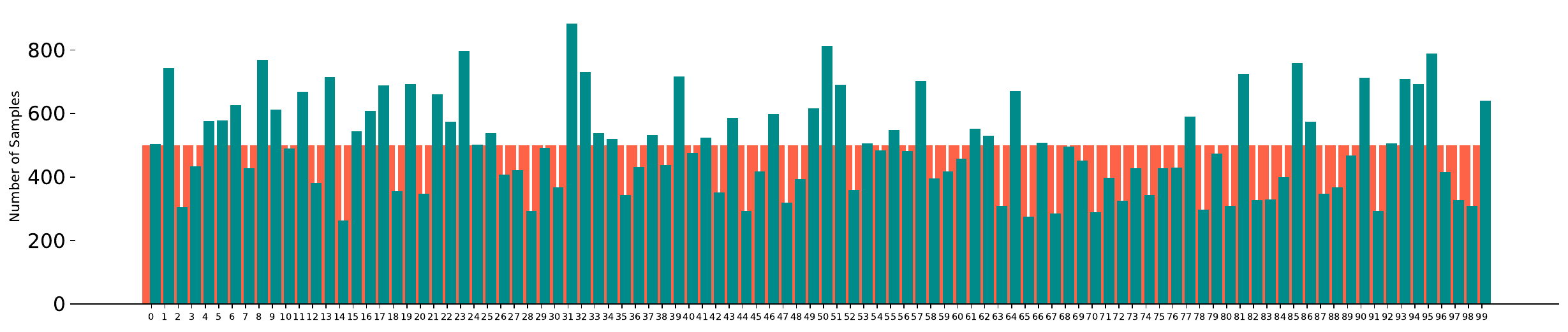}}
\vskip -0.05in
\caption{Distribution across category on CIFAR-100 of sampling with (green) $/$without (red) replacement under the Dirichlet coefficient $u=0.1$ and the number of total clients $m=100$.}
\label{distribution_with_replacement_c100}
\end{center}
\vskip -0.4in
\end{figure}

\textbf{Why we use this dataset splitting?}\\\\
Under a true federated learning framework, the distribution between data is strictly forbidden to be accessed, which means we never know the other isolated dataset aligned to other clients. In other words, it is impossible to guarantee an uniform distribution of the categories on the total dataset. Therefore, it is very valuable to explore the real performance of FL on non-uniform dataset, e.g. for the heavy tail. We also find that many algorithms that work well are significantly less effective in the face of categories imbalance of the total dataset. Thus, in our experiments, we select this difficult problem as our dataset splitting method.

\subsection{Detailed Hyperparameters Selection}
We search for a lot of hyperparameters selection to explore the best performing of the baselines and our method. Our results show that the selection of best-performing hyperparameters is mainly distinguished by two main types, vanilla local SGD and its variants like {\ttfamily FedAvg}, {\ttfamily FedCM}, {\ttfamily SCAFFOLD}, {\ttfamily FedSAM}, and the proxy-based methods like {\ttfamily FedDyn}, {\ttfamily FedSMOO}. In the Table~\ref{hyperparameters} we propose the all selections in our experiments and state the different combinations.\\
\begin{table}[h]
  \caption{Dataset introductions.}
  \vskip 0.1in
  \label{hyperparameters}
  \centering
  \begin{tabular}{ccccccc}
    \toprule
    Options     & {\ttfamily SGD}-type & Best Selection & {\ttfamily proxy}-type & Best Selection\\
    \midrule
    local learning rate   & $\left[0.01, 0.1, 0.5\right]$  & $0.1$ & $\left[0.01, 0.1, 0.5\right]$ & $0.1$    \\
    global learning rate  & $\left[0.1, 1.0\right]$ & $1.0$ & $\left[0.1, 1.0\right]$ & $1.0$    \\
    SAM learning rate     & $\left[0.001, 0.01, 0.1\right]$ & $0.01$ & $\left[0.01, 0.1, 1.0\right]$ & $0.1$    \\
    learning rate decay   & $\left[0.995, 0.998, 0.9995\right]$ & $0.998$ & $\left[0.998, 0.9995, 0.99995\right]$ & $0.9995$   \\
    penalized coefficient $\beta$ & - & - & $\left[1, 10, 100\right]$ & $10$    \\
    client-level momentum $\alpha$& $\left[0.05, 0.1, 0.5\right]$ & $0.1$ & - & -    \\
    \bottomrule
  \end{tabular}
\end{table}

For {\ttfamily FedAdam}, we use a adaptive learning rate which is different from the above due to its global adaptivity. We test a lot of selections to search for the best-performing hyperparameters.
Empirically, the global learning rate is $1.0$ for the averaged aggregation. The local learning rate usually adopts $0.1$. The SAM learning rate is $0.01$ for {\ttfamily FedSAM} and $0.1$ for {\ttfamily FedSMOO}. The learning rate decay is different from each other. Usually for the {\ttfamily SGD}-type methods, they adopt the $0.998$. While for the {\ttfamily proxy}-type methods, they require a larger selection, like $0.9995$, $0.9998$ for the stability during the training process. About the penalized coefficient $\beta$, we select $10$ for CIFAR-10 and $100$ for CIFAR-100. About the client-level momentum, we follow the \cite{FedCM} and set it as $0.1$. The other common hyperparameters are selected as introduced in the text of experiments.

\subsection{Evaluation Curves}
\begin{figure}[H]
\centering
    \subfigure[Dirichlet-0.6.]{
        \includegraphics[width=0.24\textwidth]{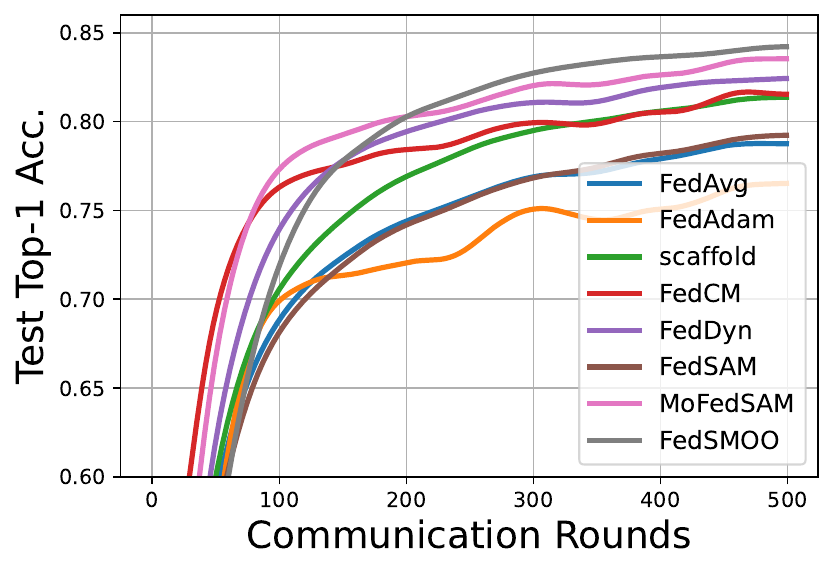}}
    \subfigure[Dirichlet-0.1.]{
	\includegraphics[width=0.24\textwidth]{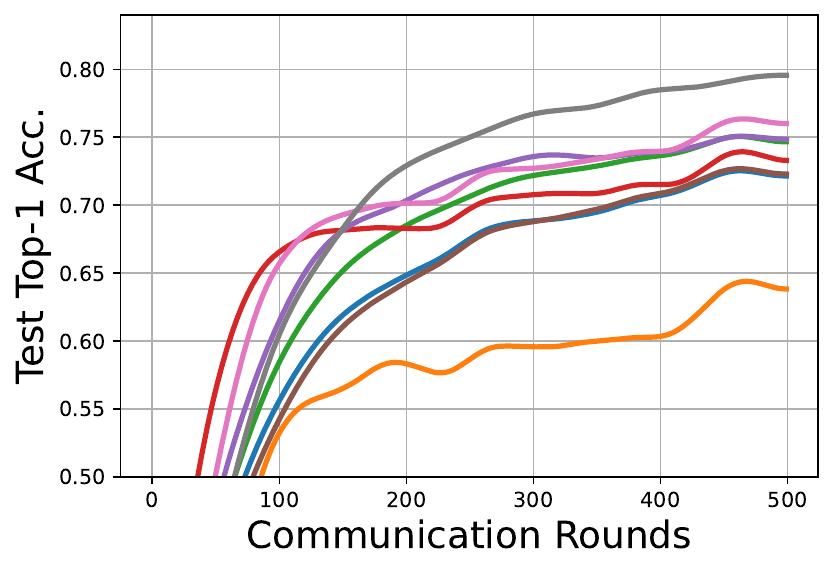}}
    \subfigure[Pathological-6.]{
	\includegraphics[width=0.24\textwidth]{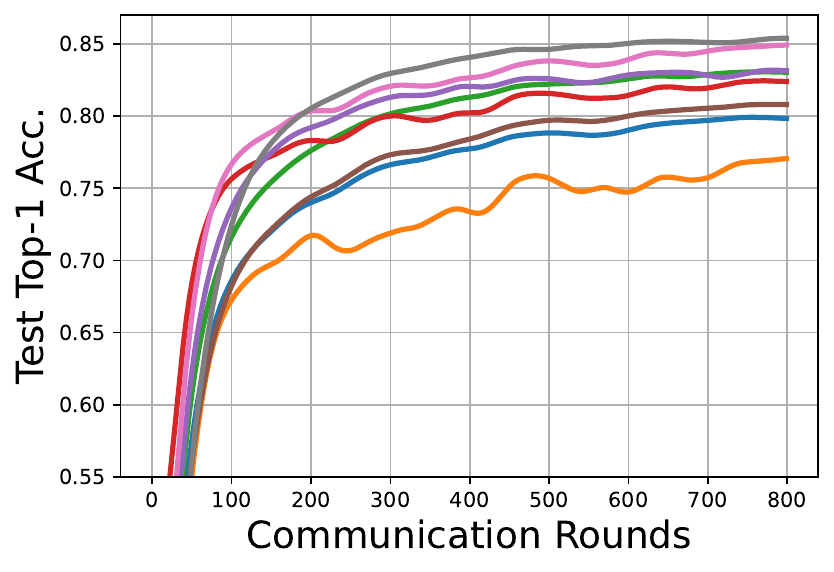}}
    \subfigure[Pathological-3.]{
	\includegraphics[width=0.24\textwidth]{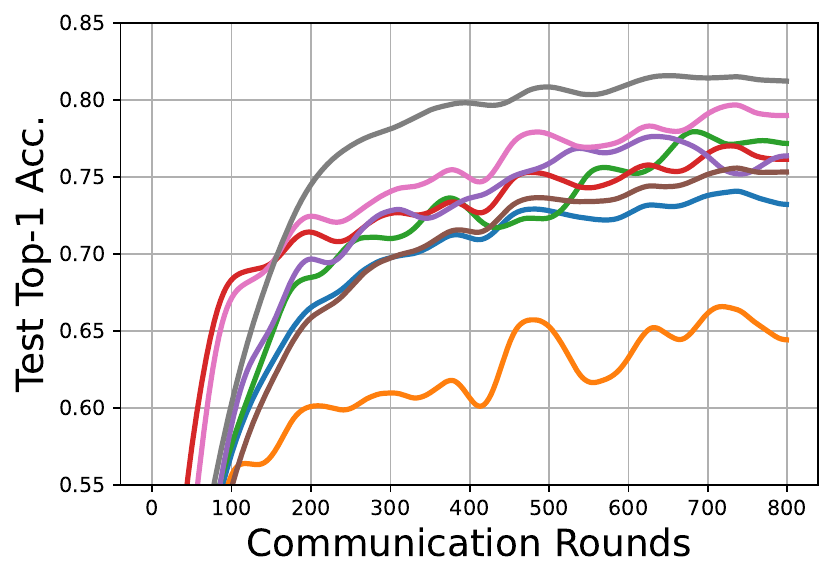}}
\vskip -0.05in
\caption{Accuracy on the CIFAR-10 dataset under $10\%$ participation of total $100$ clients.}
\label{c10-10-100-acc}
\vskip -0.05in
\end{figure}

\begin{figure}[H]
\vskip -0.2in
\centering
    \subfigure[Dirichlet-0.6.]{
        \includegraphics[width=0.24\textwidth]{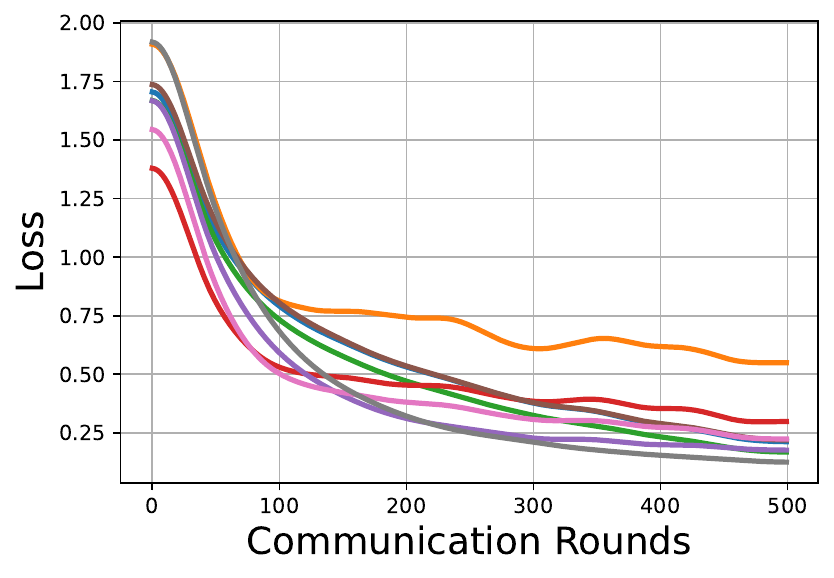}}
    \subfigure[Dirichlet-0.1.]{
	\includegraphics[width=0.24\textwidth]{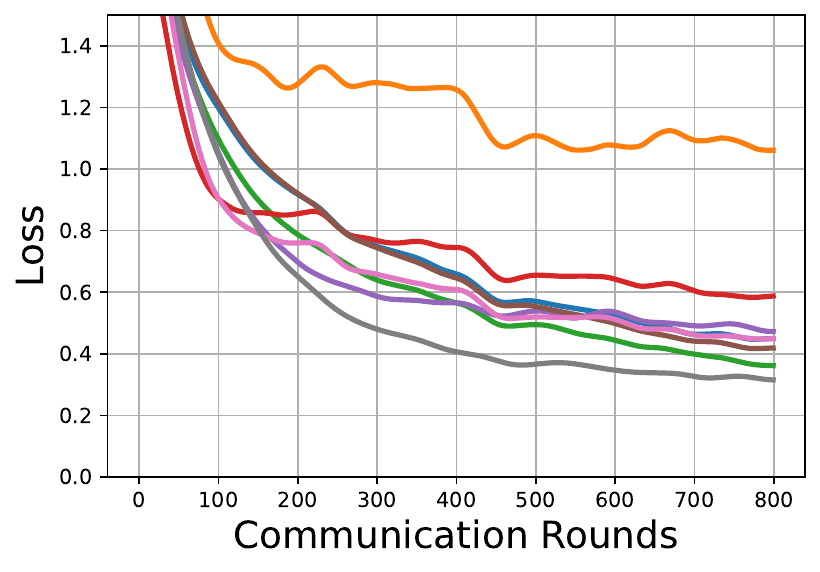}}
    \subfigure[Pathological-6.]{
	\includegraphics[width=0.24\textwidth]{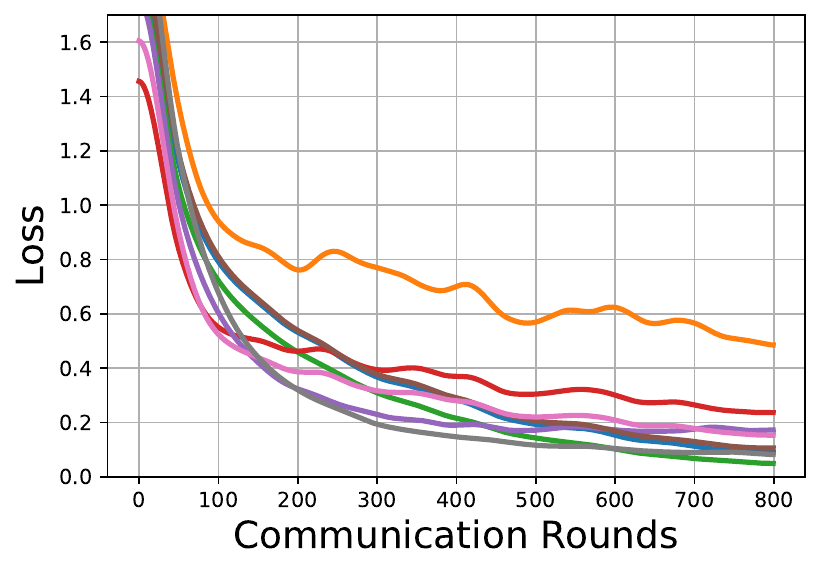}}
    \subfigure[Pathological-3.]{
	\includegraphics[width=0.24\textwidth]{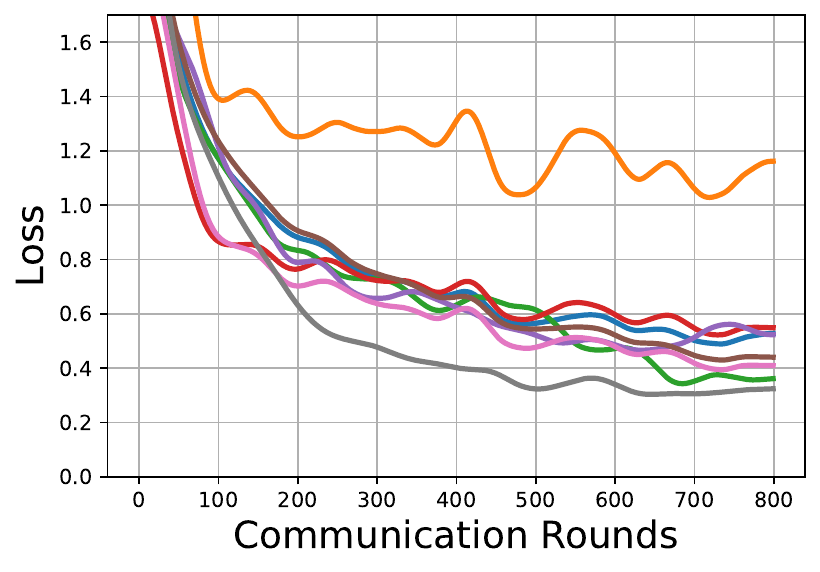}}
\vskip -0.05in
\caption{Loss on the CIFAR-10 dataset under $10\%$ participation of total $100$ clients.}
\label{c10-10-100-loss}
\vskip -0.05in
\end{figure}

\begin{figure}[H]
\vskip -0.2in
\centering
    \subfigure[Dirichlet-0.6.]{
        \includegraphics[width=0.24\textwidth]{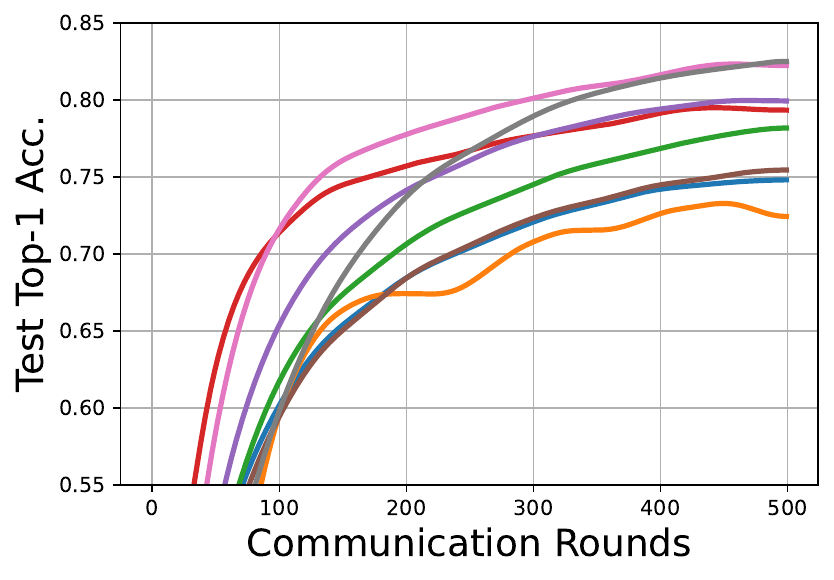}}
    \subfigure[Dirichlet-0.1.]{
	\includegraphics[width=0.24\textwidth]{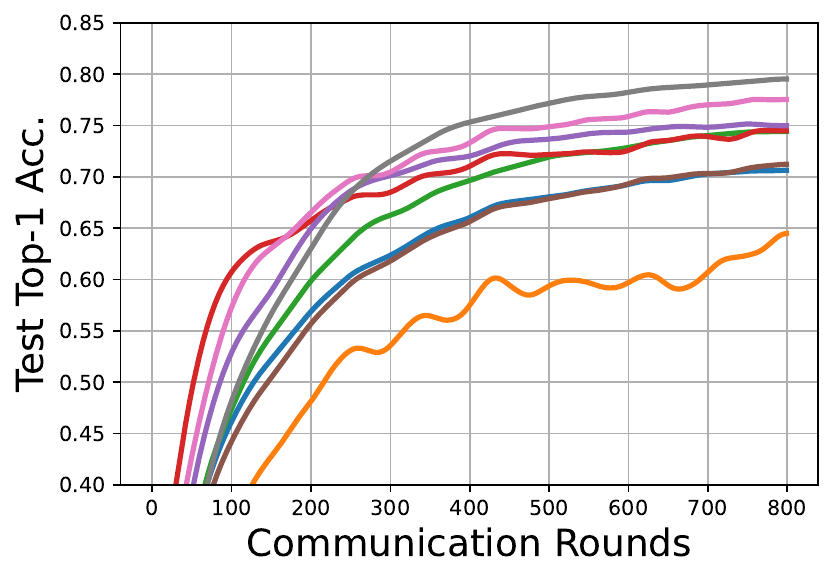}}
    \subfigure[Pathological-6.]{
	\includegraphics[width=0.24\textwidth]{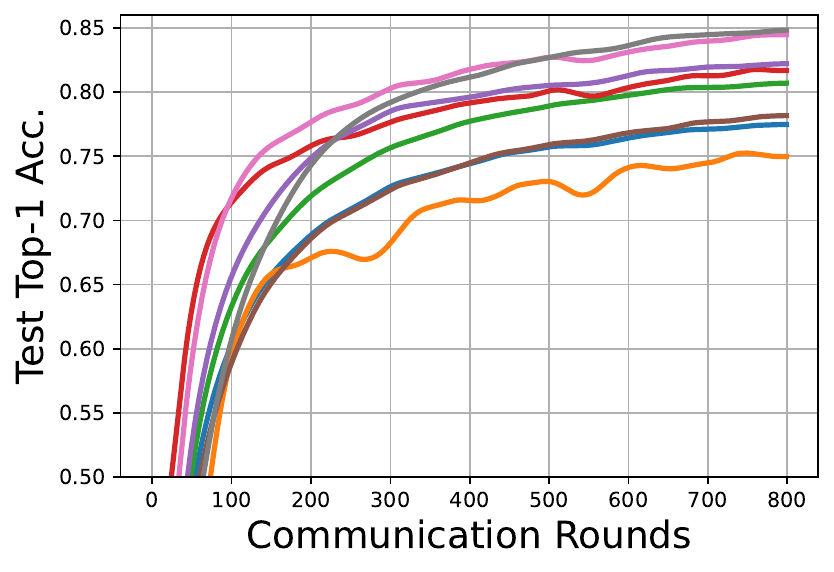}}
    \subfigure[Pathological-3.]{
	\includegraphics[width=0.24\textwidth]{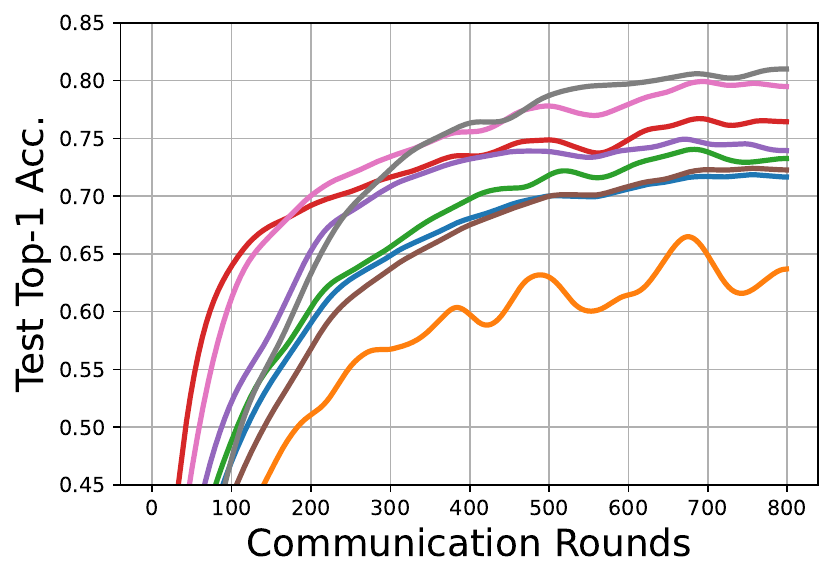}}
\vskip -0.05in
\caption{Accuracy on the CIFAR-10 dataset under $5\%$ participation of total $200$ clients.}
\label{c10-5-200-acc}
\vskip -0.05in
\end{figure}

\begin{figure}[H]
\vskip -0.2in
\centering
    \subfigure[Dirichlet-0.6.]{
        \includegraphics[width=0.24\textwidth]{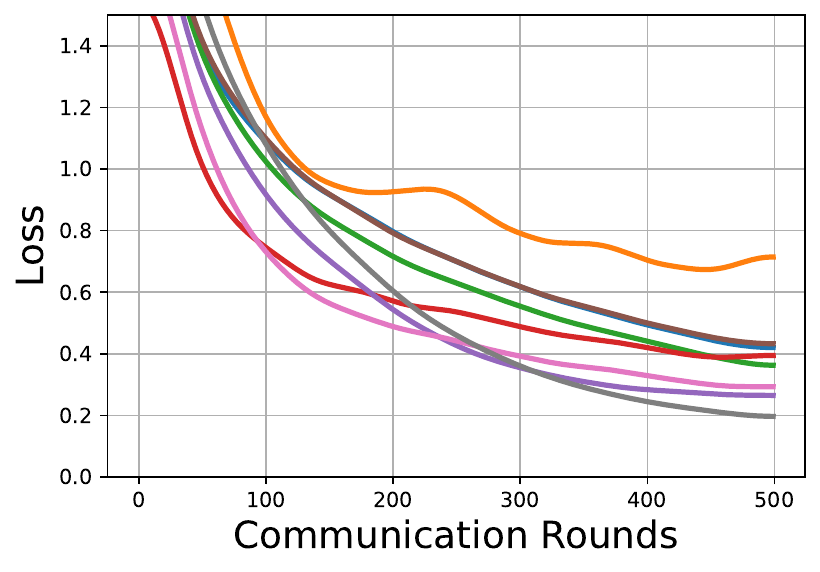}}
    \subfigure[Dirichlet-0.1.]{
	\includegraphics[width=0.24\textwidth]{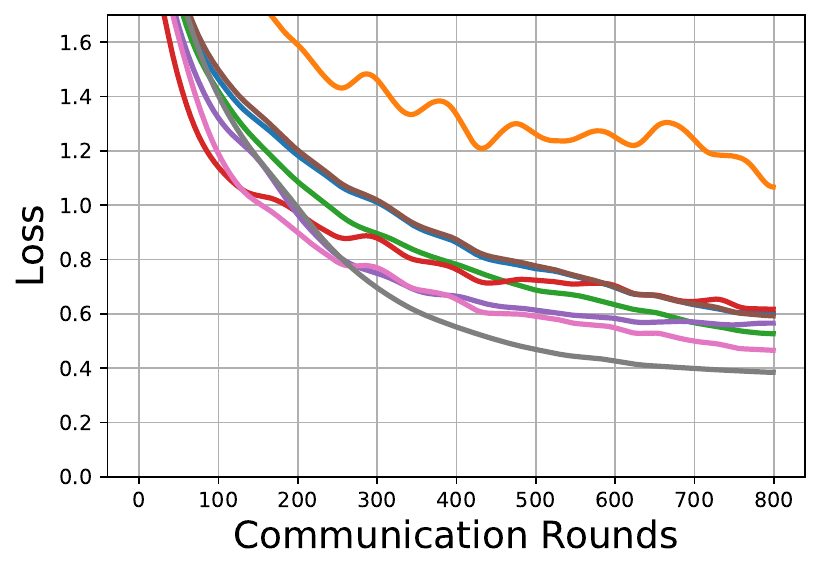}}
    \subfigure[Pathological-6.]{
	\includegraphics[width=0.24\textwidth]{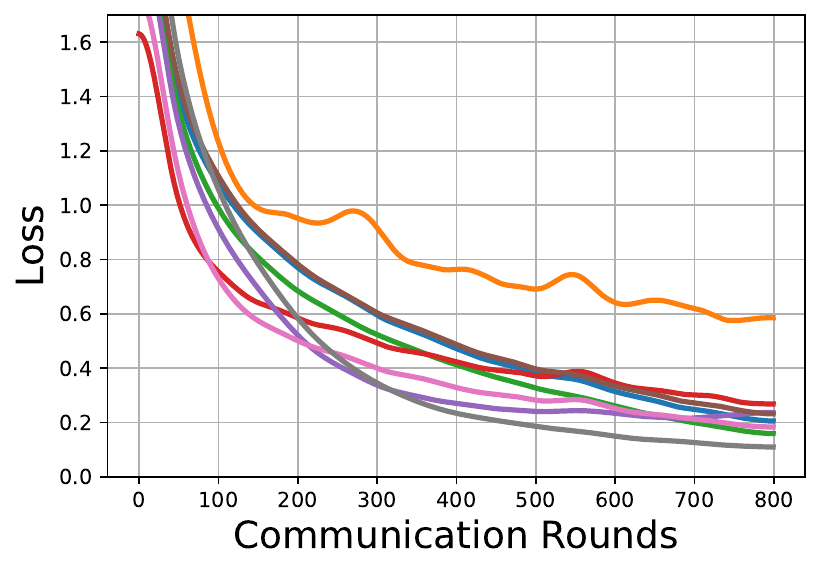}}
    \subfigure[Pathological-3.]{
	\includegraphics[width=0.24\textwidth]{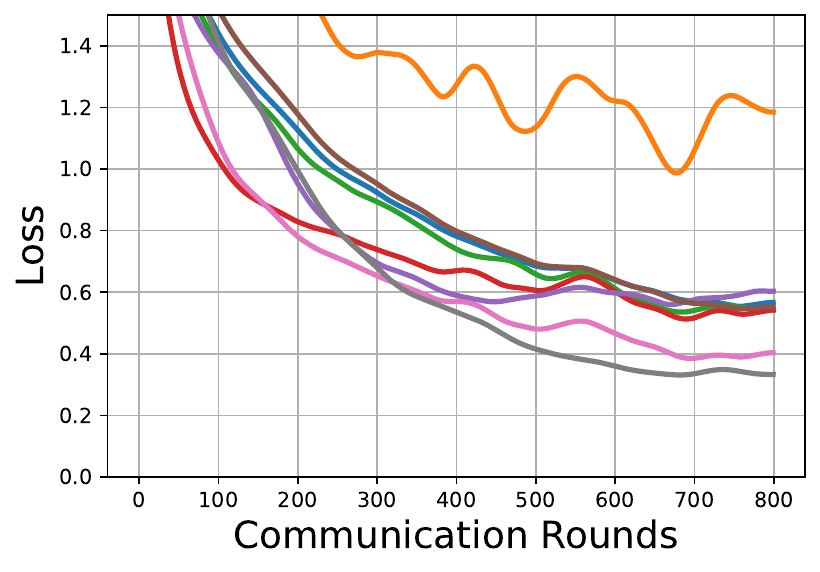}}
\vskip -0.05in
\caption{Loss on the CIFAR-10 dataset under $5\%$ participation of total $200$ clients.}
\label{c10-5-200-loss}
\vskip -0.05in
\end{figure}

To smooth the whole curve, we adopt the Hanning window as the filter to adjust them. According to the figures above, we can see that {\ttfamily FedSMOO} significantly outperforms the other algorithms, especially under the large heterogeneity, e.g. for Dirichlet-0.1 and Pathological-3. These results are in line with our expectations. Our original intention is to design the algorithm to effectively find the global flat minimum while strengthening the global consistency to obtain excellent generalization performance. When heterogeneity is low, the gaps are not large, and our algorithm remains ahead. When heterogeneity increases, our algorithm has a significant performance improvement.

\clearpage
\newpage
\subsection{Training Speed}
\begin{table}[H]
\begin{center}
\renewcommand{\arraystretch}{1}
\vspace{-0.3cm}
\caption{Communication rounds required to achieve the target test accuracy. We will note the slowest record as $1\times$ baseline to show the improvement of other algorithms on training rounds on the Dirichlet-0.1 and Pathological-10 setups. The upper part shows the results of the participation ratio equal to $10\%$-$100$ clients and the lower part shows the results of the participation ratio equal to $5\%$-$200$ clients.}
\vspace{0.15cm}
\begin{sc}
\small
\setlength{\tabcolsep}{1.95mm}{\begin{tabular}{@{}c|cccccccc@{}}
\toprule
\multicolumn{1}{c}{\multirow{3}{*}{Algorithm}} & \multicolumn{4}{c}{CIFAR-10} & \multicolumn{4}{c}{CIFAR-100} \\
\cmidrule(lr){3-4} \cmidrule(lr){7-8}
\multicolumn{1}{c}{} & \multicolumn{2}{c}{Dirichlet-0.1} & \multicolumn{2}{c}{Pathological-3} & \multicolumn{2}{c}{Dirichlet-0.1} & \multicolumn{2}{c}{Pathological-10} \\ 
\cmidrule(lr){2-5} \cmidrule(lr){6-9}
\multicolumn{1}{c}{} & \multicolumn{1}{c}{$acc=68\%$} & \multicolumn{1}{c}{$acc=74\%$} & \multicolumn{1}{c}{$acc=68\%$} & \multicolumn{1}{c}{$acc=74\%$} & \multicolumn{1}{c}{$acc=38\%$} & \multicolumn{1}{c}{$acc=43\%$} & \multicolumn{1}{c}{$acc=35\%$} & \multicolumn{1}{c}{$acc=40\%$} \\
\cmidrule(lr){1-1} \cmidrule(lr){2-9}
{\ttfamily FedAvg}       & 259 &  -  & 246 & 723 & 339 & 744 & 373 & 732 \\ 
{\ttfamily FedAdam}      & 445 &  -  & 487 &  -  & 772 &  -  & 788 &  -  \\ 
{\ttfamily SCAFFOLD}     & 191 & 419 & 180 & 533 & 323 & 480 & 361 & 617 \\ 
{\ttfamily FedCM}        & 133 & 620 &  96 & 442 & 264 & 603 & 447 &  -  \\ 
{\ttfamily FedDyn}       & 148 & 410 & 173 & 424 & 239 &  -  & 267 &  -  \\ 
{\ttfamily FedSAM}       & 264 & 649 & 253 & 604 & 361 & 603 & 380 & 718 \\ 
{\ttfamily MoFedSAM}     & 122 & 404 & 116 & 298 & 271 & 445 & 450 & 599 \\ 
{\ttfamily \textbf{Our}} & 132 & 224 & 144 & 194 & 252 & 350 & 258 & 354 \\ 
\cmidrule(lr){1-1} \cmidrule(lr){2-5} \cmidrule(lr){6-9}
\multicolumn{1}{c}{} & \multicolumn{1}{c}{$acc=68\%$} & \multicolumn{1}{c}{$acc=74\%$} & \multicolumn{1}{c}{$acc=68\%$} & \multicolumn{1}{c}{$acc=74\%$} & \multicolumn{1}{c}{$acc=35\%$} & \multicolumn{1}{c}{$acc=40\%$} & \multicolumn{1}{c}{$acc=33\%$} & \multicolumn{1}{c}{$acc=38\%$} \\
\cmidrule(lr){1-1} \cmidrule(lr){2-9}
{\ttfamily FedAvg}       & 494 &  -  & 397 &  -  & 438 & 747 & 529 &  -  \\ 
{\ttfamily FedAdam}      & 688 &  -  &  -  &  -  & 700 &  -  &  -  &  -  \\ 
{\ttfamily SCAFFOLD}     & 345 & 696 & 352 & 674 & 433 & 609 & 518 &  -  \\ 
{\ttfamily FedCM}        & 251 & 744 & 167 & 438 & 255 & 437 & 374 & 751 \\ 
{\ttfamily FedDyn}       & 239 & 532 & 234 & 599 & 323 & 600 & 418 & 750 \\ 
{\ttfamily FedSAM}       & 509 &  -  & 419 &  -  & 478 & 729 & 556 &  -  \\ 
{\ttfamily MoFedSAM}     & 220 & 415 & 170 & 355 & 298 & 432 & 412 & 621 \\ 
{\ttfamily \textbf{Our}} & 245 & 355 & 239 & 330 & 352 & 534 & 447 & 541 \\ 
\bottomrule
\end{tabular}}
\label{speed}
\end{sc}
\end{center}
\vspace{-0.3cm}
\end{table}
According to the above Table, we can see that our method performs well on average. To achieve the high accuracy, it is much faster than other benchmarks. SAM optimizers usually make the entire training process slower, see results of {\ttfamily FedAvg} and {\ttfamily FedSAM}. However, enhancing consistency will improve this. {\ttfamily MoFedSAM} force the consistency by adopting a global momentum on each local client weighted by a coefficient $\alpha$~(usually 0.1), which means in the local updates of round $t$, their gradients share a common direction as $90\%$ and their own local gradients as $10\%$. {\ttfamily FedSMOO} provides a amended vector on both parameters $w$ and perturbation $s$, which greatly enhances its consistency. Thus, our method can efficiently train the federated model, especially on the large heterogeneity.

\begin{table}[H]
\label{time}
\small
    \caption{Training wall-clock time comparison to achieve $74\%$ accuracy on the CIFAR-10 of Pathological-3 dataset split.}
    \vspace{0.1cm}
    \label{fig:different_selection}
    \centering
    \begin{tabular}{ccccc}
   \toprule
     $\alpha_{1}$ & Times~(s/Round) & Rounds & Total Time~(s)  \\
    \midrule
     {\ttfamily FedAvg}      & 13.33~($1.76\times$) & 723~($1\times$)     &  9637.59~($1\times$)   \\
     {\ttfamily FedAdam}     & 13.77~($1.70\times$) &  -  &  -  \\
     {\ttfamily SCAFFOLD}    & 16.97~($1.38\times$) & 533~($1.36\times$)  &  9045.01~($1.06\times$)   \\
     {\ttfamily FedCM}       & 14.36~($1.63\times$) & 442~($1.64\times$)  &  6347.12~($1.51\times$)   \\
     {\ttfamily FedDyn}      & 12.81~($1.83\times$) & 424~($1.71\times$)  &  5431.44~($1.77\times$)   \\
     {\ttfamily FedSAM}      & 19.69~($1.19\times$) & 604~($1.20\times$)  & 11892.76~($0.81\times$)   \\
     {\ttfamily MoFedSAM}    & 22.08~($1.06\times$) & 298~($2.43\times$)  &  6579.84~($1.46\times$)   \\
     \textbf{{\ttfamily\textbf{OUR}}} & \textbf{23.52~($1\times$)} & \textbf{194~($3.73\times$)}     &  \textbf{4562.88~($2.11\times$)}   \\
    \bottomrule
\end{tabular}
\vspace{-0.2cm}
\end{table}
\textbf{Test Experiments:} A100-SXM4-40GB GPU, CUDA Driver 11.7, Driver Version 515.86.01, PyTorch-1.13.1 

Table~\ref{time} shows the wall-clock time costs on the CIFAR-10 of the Pathological-3 dataset split. Due to the double calculation of the gradients via SAM optimizer, {\ttfamily FedSAM}, {\ttfamily MoFedSAM}, and {\ttfamily FedSMOO} will take more time in a single round of updates, about $1.5\times$ over the single-calculation methods. However, the communication rounds required are much less than the single-calculation methods. Specifically, to achieve the $74\%$ accuracy, {\ttfamily FedSMOO} can be accelerated by a factor of $3.73$ on the communication rounds of the {\ttfamily FedAvg}. Considering the total wall-clock time costs, the acceleration ratio achieves $2.11\times$. An important case is that {\ttfamily FedSAM} can not accelerate the total wall-clock time due to the double-calculation, which must be combined with other methods to further improve its performance, like its variant {\ttfamily MoFedSAM}.

\clearpage
\newpage
\subsection{Ablation studies}
\begin{table}[H]\label{abla}
\small
    \caption{Ablation studies of different modules.}
    \vspace{0.1cm}
    \centering
    \begin{tabular}{cccc}
   \toprule
     Dynamic Regularization & SAM & SAM-correction & Accuracy ($\%$)\\
    \midrule
     - & - & - &  $74.08_{\pm.22}$ \\
     $\sqrt{}$ & - & - &  $77.63_{\pm.14}$ \\
     $\sqrt{}$ & $\sqrt{}$ & - & $80.13_{\pm.11}$  \\
     $\sqrt{}$ & $\sqrt{}$ & $\sqrt{}$ &  $81.58_{\pm.16}$ \\
    \bottomrule
\end{tabular}
\end{table}
We test the performance of the different modules on the CIFAR-10 dataset of Pathological-3 split, which are named as "Dynamic Regularization", "SAM", and "SAM-correction" modules. The vanilla benchmark is {\ttfamily FedAvg}. After introducing the above three modules, the performance improvements are $3.55\%$, $6.05\%$, and $7.5\%$.

\clearpage
\newpage
\section{Proofs}
\label{appendix:proofs}

\subsection{Proof of Theorem 4.1}
\subsubsection{Preliminary Lemmas}
Before proving the theorem, we first introduce some preliminary lemmas used in our proofs.
\begin{lemma}
    For random variables $\{x_k\}_{k\in[K]}\in\mathbb{R}^d$, we can bound:
    \begin{equation}
        \mathbb{E}\Vert\sum_k x_k\Vert^2\leq K\sum_k\mathbb{E}\Vert x_k\Vert^2.
    \end{equation}
    \begin{proof}
        It can be proved by Jensen inequality.
    \end{proof}
\end{lemma}

\begin{lemma}
    For two random variables $x,y\in\mathbb{R}^d$, we can bound:
    \begin{equation}
        \Vert x+y\Vert^2\leq\left(1+\frac{1}{c}\right)\Vert x\Vert^2 + \left(1+c\right) \Vert y\Vert^2,
    \end{equation}
    where $c > 0$ is a constant.
    \begin{proof}
        It can be proved by triangle inequality.
    \end{proof}
\end{lemma}

\begin{lemma}
    For two random variables $x,y\in\mathbb{R}^d$, we can bound:
    \begin{equation}
        \langle x,y\rangle \leq \frac{1}{2}\Vert x \Vert^2 - \frac{1}{2}\Vert x - y\Vert^2.
    \end{equation}
    \begin{proof}
        Applying the product of two vectors:
        \begin{align*}
            \Vert x - y\Vert^2 
            &= \langle x-y,x-y\rangle=\langle x,x\rangle+\langle x,-y\rangle+\langle -y,x\rangle+\langle -y,-y\rangle= \Vert x \Vert^2 + \Vert y \Vert^2 - 2\langle x,y\rangle.\\
        \end{align*}
        Thus we can bound the product term as:
        \begin{align*}
            \langle x,y\rangle = \frac{1}{2}\Vert x \Vert^2 + \frac{1}{2}\Vert y \Vert^2 - \frac{1}{2}\Vert x - y\Vert^2\leq \frac{1}{2}\Vert x \Vert^2 - \frac{1}{2}\Vert x - y\Vert^2.\\
        \end{align*}
    \end{proof}
\end{lemma}

\subsubsection{Proofs of Theorem 4.1}
The paradigm of our proposed method is shown in Algorithm~\ref{algorithm}. We use a similar proof technique in \cite{SCAFFOLD,FedDyn} and define a set of auxiliary variables to match the global update. With the partial participation training, the active clients at each communication round will update the local parameters while the inactive clients will inherit the previous parameters. Thus, we introduce the virtual variables $\widetilde{w}$ as:
\begin{equation}
    \widetilde{w}_{i}^{t} = \mathop{\arg\min}\limits_{w} \left\{\mathcal{F}_i(w) - \langle \lambda_i^t,w-w^{t}\rangle + \frac{1}{2\beta}\Vert w - w^{t}\Vert^{2}\right\}, \quad i\in\left[\mathcal{m}\right].
\end{equation}
The virtual variable is based on partial participation. The active client set is randomly selected with the probability of $\frac{n}{m}$, which means that $w_i^t$ equals to $\widetilde{w}_i^t$ with probability $\frac{n}{m}$ and maintains $\widetilde{w}_i^{t-1}$ otherwise. Thus the first order condition satisfies: $\beta\left(\nabla\mathcal{F}_i(\widetilde{w}_{i}^{t}) -\lambda_i^t\right) + \widetilde{w}_{i}^{t}-w^{t}=0$, which indicates that $\lambda_i^{t+1} = \nabla\mathcal{F}_i(w_{i}^{t})$ among active $i\in \left[\mathcal{n}\right]$ after the local update of the dual variable. If the client $i$ is inactive, the $\widetilde{w}_{i}^{t}$ maintains the previous parameters $w^t$. Thus the global variable $\lambda$ is updated as:
\begin{align*}
    \lambda^{t+1} - \lambda^t 
    &= - \frac{1}{\beta m}\sum_{i\in \left[\mathcal{n}\right]}\left(w_{i}^{t}-w^t\right) = - \frac{1}{\beta m}\sum_{i\in \left[\mathcal{m}\right]}\left(\widetilde{w}_{i}^{t}-w^t\right) \\
    &= \frac{1}{\beta m}\sum_{i\in\left[\mathcal{m}\right]}\left(\nabla\mathcal{F}_i(\widetilde{w}_{i}^{t}) - \lambda_i^t\right) = \frac{1}{\beta m}\sum_{i\in \left[\mathcal{m}\right]}(\lambda_i^{t+1}-\lambda_i^t).\\
\end{align*}
According to the update of the $\lambda$, we consider that $\lambda^t=\frac{1}{m}\sum_{i\in \left[\mathcal{m}\right]}\lambda_i^t$ as $\lambda_i^{t+1}=\lambda_i^t$ holds for $i\in \left[\mathcal{m}\right] - \left[\mathcal{n}\right]$. In order to distinguish the parameters before and after updating with the dual variable, we define the:
\begin{equation}
\label{update_of_overline_w}
    \overline{w}^{t+1}\triangleq\frac{1}{n}\sum_{i\in \left[\mathcal{n}\right]}w_{i}^t=w^{t+1}+\beta\lambda^{t+1}.
\end{equation}
\begin{lemma}
\rm
\label{l1}
    The adjacent averaged parameters generated in Algorithm~\ref{algorithm} satisfies the following update:
    \begin{equation}
        \mathbb{E}\left[\overline{w}^{t+1}\right]-\mathbb{E}\left[\overline{w}^{t}\right] = -\frac{\beta}{m}\sum_{i\in \left[\mathcal{m}\right]}\mathbb{E}\left[\nabla \mathcal{F}_i(\widetilde{w}_{i}^{t}) \right].
    \end{equation}
    \begin{proof}
    \rm
    According to the equation~{(\ref{update_of_overline_w})}, we have:
        \begin{align*}
            \mathbb{E}\left[\overline{w}^{t+1}\right]-\mathbb{E}\left[\overline{w}^{t}\right]
            &= \mathbb{E}\left[\frac{1}{n}\sum_{i\in \left[\mathcal{n}\right]}\left(w_{i}^t-w^t-\beta\lambda^t\right)\right] = \mathbb{E}\left[\frac{1}{m}\sum_{i\in \left[\mathcal{m}\right]}\left(\widetilde{w}_{i}^t-w^t-\beta\lambda^t\right)\right]\\
            &= \frac{\beta}{m}\sum_{i\in \left[\mathcal{m}\right]}\mathbb{E}\left[\lambda_i^t-\lambda^t-\nabla\mathcal{F}_i(\widetilde{w}_{i}^{t})\right] = -\frac{\beta}{m}\sum_{i\in \left[\mathcal{m}\right]}\mathbb{E}\left[\nabla \mathcal{F}_i(\widetilde{w}_{i}^{t}) \right].\\
        \end{align*}
    \end{proof}
    The second equation is based on the expectation of $\widetilde{w}$ equal to the expectation of clients' sampling. The last equation is based on the relationship between $\lambda_i$ and $\lambda$.
\end{lemma}
According to the lemma~\ref{l1}, we can directly bound the norm term of the update of $\overline{w}$:
\begin{equation}
\label{bound_of_overline_w}
    \mathbb{E}\Vert \overline{w}^{t+1}-\overline{w}^t\Vert^{2}=\mathbb{E}\Vert \frac{1}{n}\sum_{i\in\left[\mathcal{n}\right]}w_i^t-\overline{w}^t\Vert^{2}\leq \frac{1}{n}\mathbb{E}\sum_{i\in\left[\mathcal{n}\right]}\Vert w_i^{t}-\overline{w}^t\Vert^{2}\leq \frac{1}{m}\sum_{i\in\left[\mathcal{m}\right]}\mathbb{E}\Vert \widetilde{w}_i^{t}-\overline{w}^t\Vert^{2}.
\end{equation}
The last inequality is based on the expectation of $\widetilde{w}_i$ equals to the expectation of clients' sampling.\\
In the inequality~(\ref{bound_of_overline_w}), the term $\widetilde{w}_i$ is the local parameters generated in algorithm~\ref{algorithm} at each round $t$, and the $\overline{w}^t$ is the averaged parameters. Thus the \textit{RHS} term of the inequality~(\ref{bound_of_overline_w}) represents the average norm expectation of the local changes before dual correction. It bounds that the norm of the global update will not exceed its average local updates. In this proof, we denote $\upsilon^t\triangleq\frac{1}{m}\sum_{i\in\left[\mathcal{m}\right]}\mathbb{E}\Vert \widetilde{w}_i^{t}-\overline{w}^t\Vert^{2}$ and provide its upper bound in the next part. Furthermore, we need to consider another important term $\frac{1}{m}\sum_{i\in\left[\mathcal{m}\right]}\mathbb{E}\Vert w_i^{t}-\overline{w}^{t+1}\Vert^{2}$, the averaged divergence of the heterogeneity during the local training, which is also defined as the 'client drift' in FL. It reflects the volatility caused by the heterogeneity in the local training process. In this proof, we denote $\delta^t$ as the divergence term at round $t$. In the next part, we will bound these two terms to demonstrate the convergence analysis.

\begin{lemma}
\rm
\label{l2}
Based on the assumptions, the averaged local update term could be bounded as:
\begin{equation}
    (1 - 12\beta^2L^2)\upsilon^t \leq 30\beta^2L^2r^2 + 36\beta^2L^2\delta^{t-1} + 12\beta^2\mathbb{E}\Vert \nabla f(\overline{w}^t)\Vert^2,
\end{equation}
where $\beta < \frac{1}{2\sqrt{3}L}$ for $1-12\beta^2L^2 > 0$.
\begin{proof}
\rm
According to the update rules~(\ref{update_of_overline_w}) and the first order gradient condition above, the $\upsilon^t$ term could be bounded as:
    \begin{align*}
        \upsilon^t
        &= \frac{1}{m}\sum_{i\in\left[\mathcal{m}\right]}\mathbb{E}\Vert \widetilde{w}_i^{t}-\overline{w}^t\Vert^{2}
         = \frac{1}{m}\sum_{i\in\left[\mathcal{m}\right]}\mathbb{E}\Vert \widetilde{w}_i^{t}-w^{t}-\beta\lambda^{t}\Vert^{2}\\
        &= \frac{\beta^2}{m}\sum_{i\in\left[\mathcal{m}\right]}\mathbb{E}\Vert \lambda_i^t -\nabla\mathcal{F}_i(\widetilde{w}_i^t)-\lambda^{t}\Vert^{2} = \frac{\beta^2}{m}\sum_{i\in\left[\mathcal{m}\right]}\mathbb{E}\Vert \nabla f_i(w_i^{t-1}+\hat{s}_i^{t-1}) -\nabla f_i(\widetilde{w}_i^t+\hat{s}_i^t)-\lambda^{t}\Vert^{2}\\
        &= \frac{\beta^2}{m}\sum_{i\in\left[\mathcal{m}\right]}\mathbb{E}\Vert \nabla f_i(w_i^{t-1}+\hat{s}_i^{t-1}) - \nabla f_i(w_i^{t-1}) + \nabla f_i(w_i^{t-1}) - \nabla f_i(\widetilde{w}_i^t+\hat{s}_i^t) + \nabla f_i(\widetilde{w}_i^t) - \nabla f_i(\widetilde{w}_i^t) -\lambda^{t}\Vert^{2}\\
        &\leq \frac{3\beta^2}{m}\sum_{i\in\left[\mathcal{m}\right]}\mathbb{E}\Vert \nabla f_i(w_i^{t-1}+\hat{s}_i^{t-1}) - \nabla f_i(w_i^{t-1})\Vert^2 + \frac{3\beta^2}{m}\sum_{i\in\left[\mathcal{m}\right]}\mathbb{E}\Vert\nabla f_i(\widetilde{w}_i^t+\hat{s}_i^t) - \nabla f_i(\widetilde{w}_i^t)\Vert^2\\
        &\quad + \frac{3\beta^2}{m}\sum_{i\in\left[\mathcal{m}\right]}\mathbb{E}\Vert \nabla f_i(w_i^{t-1}) - \nabla f_i(\widetilde{w}_i^t) -\lambda^{t}\Vert^{2}\\
        &\leq \frac{3\beta^2L^2}{m}\sum_{i\in\left[\mathcal{m}\right]}\Vert \hat{s}_i^{t-1}\Vert^2 + \frac{3\beta^2L^2}{m}\sum_{i\in\left[\mathcal{m}\right]}\Vert \hat{s}_i^{t}\Vert^2 + \frac{3\beta^2}{m}\sum_{i\in\left[\mathcal{m}\right]}\mathbb{E}\Vert \nabla f_i(w_i^{t-1}) - \nabla f_i(\widetilde{w}_i^t) -\lambda^{t}\Vert^{2}\\
        &= \frac{3\beta^2L^2}{m}\sum_{i\in\left[\mathcal{m}\right]}\Vert r\frac{ \mathcal{s}_i^{t-1}}{\Vert \mathcal{s}_i^{t-1}\Vert}\Vert^2 + \frac{3\beta^2L^2}{m}\sum_{i\in\left[\mathcal{m}\right]}\Vert r\frac{ \mathcal{s}_i^{t}}{\Vert \mathcal{s}_i^{t}\Vert}\Vert^2 + \frac{3\beta^2}{m}\sum_{i\in\left[\mathcal{m}\right]}\mathbb{E}\Vert \nabla f_i(w_i^{t-1}) - \nabla f_i(\widetilde{w}_i^t) -\lambda^{t}\Vert^{2}\\
        &= 6\beta^2L^2r^2 + \frac{3\beta^2}{m}\sum_{i\in\left[\mathcal{m}\right]}\mathbb{E}\Vert \nabla f_i(w_i^{t-1}) - \nabla f_i(\widetilde{w}_i^t) -\lambda^{t}\Vert^{2}\\
        &= 6\beta^2L^2r^2 + \frac{3\beta^2}{m}\sum_{i\in\left[\mathcal{m}\right]}\mathbb{E}\Vert \nabla f_i(w_i^{t-1}) - \nabla f_i(\overline{w}^t) + \nabla f_i(\overline{w}^t) - \nabla f_i(\widetilde{w}_i^t) - \nabla f(\overline{w}^t) + \nabla f(\overline{w}^t) -\lambda^{t}\Vert^{2}\\
        &\leq 6\beta^2L^2r^2 + \frac{12\beta^2L^2}{m}\sum_{i\in\left[\mathcal{m}\right]}\mathbb{E}\Vert w_i^{t-1} - \overline{w}^t\Vert^2 + \frac{12\beta^2L^2}{m}\sum_{i\in\left[\mathcal{m}\right]}\mathbb{E}\Vert\overline{w}^t - \widetilde{w}_i^t\Vert^2 + 12\beta^2\mathbb{E}\Vert \nabla f(\overline{w}^t)\Vert^2\\
        &\quad + 12\beta^2\mathbb{E}\Vert \frac{1}{m}\sum_{i\in\left[\mathcal{m}\right]} \left( \nabla f_i(\overline{w}^t) -\lambda_i^{t}\right)\Vert^2 \\
        &\leq 6\beta^2L^2r^2 + 12\beta^2L^2\delta^{t-1} + 12\beta^2L^2\upsilon^t + 12\beta^2\mathbb{E}\Vert \nabla f(\overline{w}^t)\Vert^2 + \frac{12\beta^2}{m}\sum_{i\in\left[\mathcal{m}\right]}\mathbb{E}\Vert \nabla f_i(\overline{w}^t) -\lambda_i^{t}\Vert^2\\
        &\leq 6\beta^2L^2r^2 + 12\beta^2L^2\delta^{t-1} + 12\beta^2L^2\upsilon^t + 12\beta^2\mathbb{E}\Vert \nabla f(\overline{w}^t)\Vert^2\\
        &\quad + \frac{12\beta^2}{m}\sum_{i\in\left[\mathcal{m}\right]}\mathbb{E}\Vert \nabla f_i(\overline{w}^t) - \nabla f_i(w_i^{t-1}) + \nabla f_i(w_i^{t-1}) -\nabla f_i(w_i^{t-1}+\hat{s}_i^{t-1})\Vert^2\\
        &\leq 30\beta^2L^2r^2 + 36\beta^2L^2\delta^{t-1} + 12\beta^2L^2\upsilon^t + 12\beta^2\mathbb{E}\Vert \nabla f(\overline{w}^t)\Vert^2.\\
    \end{align*}
Collecting the like terms of \textit{LHS} and \textit{RHS} above will complete the proofs.
\end{proof}
\end{lemma}
Lemma~\ref{l2} reveal one relationship between the local updates and the divergence drifts. When we provide a constant bounded $r$, the upper bound of the local updates is always controlled by the global gradient term and divergence drifts, which is consistent with the conclusion of the gradient descent method in FL. In order to further clarify the relationship between these two terms, we introduce the upper bound of divergence drift term $\delta$ in the next.
\begin{lemma}
\label{l3}
\rm
Based on the assumptions, the divergence drifts term could be bounded as:
\begin{equation}
    \delta^t \leq \frac{2m-2n}{n}\delta^{t-1} + \frac{6m-2n}{2m-n}\upsilon^t.
\end{equation}
\begin{proof}
\rm
According to the definition of $delta$, we have:
\begin{align*}
    \delta^t
    &=\frac{1}{m}\sum_{i\in\left[\mathcal{m}\right]}\mathbb{E}\Vert w_i^{t}-\overline{w}^{t+1}\Vert^{2}
     =\frac{1}{m}\sum_{i\in\left[\mathcal{m}\right]}\mathbb{E}\Vert w_i^{t}-\overline{w}^{t}+\overline{w}^{t}-\overline{w}^{t+1}\Vert^{2}\\
    &\leq \left(1+\frac{1}{c}\right)\frac{1}{m}\sum_{i\in\left[\mathcal{m}\right]}\mathbb{E}\Vert w_i^{t}-\overline{w}^{t}\Vert^{2} + \left(1+c\right)\mathbb{E}\Vert \overline{w}^{t}-\overline{w}^{t+1}\Vert^{2}\\
    &= \frac{n}{m}\left(1+\frac{1}{c}\right)\frac{1}{m}\sum_{i\in\left[\mathcal{m}\right]}\mathbb{E}\Vert \widetilde{w}_i^{t}-\overline{w}^{t}\Vert^{2} + \frac{m-n}{m}(1+\frac{1}{c})\frac{1}{m}\sum_{i\in\left[\mathcal{m}\right]}\mathbb{E}\Vert w_i^{t-1}-\overline{w}^{t}\Vert^{2} + \left(1+c\right)\mathbb{E}\Vert \overline{w}^{t}-\overline{w}^{t+1}\Vert^{2}\\
    &= \left[\frac{n}{m}\left(1+\frac{1}{c}\right)+(1+c)\right]\upsilon^t + \frac{m-n}{m}\left(1+\frac{1}{c}\right)\delta^{t-1}.\\
\end{align*}
\end{proof}
For the convenience of analysis, we select a proper constant $c$ to complete the proofs. Let $c=\frac{n}{2m-n} > 0 $ where $n < m$, thus we have $1 + c = \frac{2m}{2m-n}$ and $1+\frac{1}{c}=\frac{2m}{n}$. Plugging into the last equation and we have: $\left[\frac{n}{m}\left(1+\frac{1}{c}\right)+(1+c)\right] = \frac{6m-2n}{2m-n}$ and $\frac{m-n}{m}\left(1+\frac{1}{c}\right)=\frac{2m-2n}{n}$, which completes the proofs.
\end{lemma}
According to the lemmas above, we can bound the recursion terms of $\upsilon^t$ and $\delta^t$ to prove the convergence theorem. From the assumption of smoothness, taking the conditional expectation with $t$ on both side and we have:
\begin{align*}
    &\quad \mathbb{E}_t\left[f(\overline{w}^{t+1})\right] \\
    &\leq f(\overline{w}^t) + \frac{L}{2}\mathbb{E}_t\Vert\overline{w}^{t+1}-\overline{w}^t\Vert^2 + \mathbb{E}_t\langle\nabla f(\overline{w}^t),\overline{w}^{t+1}-\overline{w}^t\rangle \\
    &= f(\overline{w}^t) + \frac{L}{2}\mathbb{E}_t\Vert\overline{w}^{t+1}-\overline{w}^t\Vert^2 + \mathbb{E}_t\langle\nabla f(\overline{w}^t),-\frac{\beta}{m}\sum_{i\in \left[\mathcal{m}\right]}\mathbb{E}_t\left[\nabla \mathcal{F}_i(\widetilde{w}_{i}^{t}) \right]\rangle\\
    &\leq f(\overline{w}^t) + \frac{L}{2}\mathbb{E}_t\Vert\overline{w}^{t+1}-\overline{w}^t\Vert^2 + \frac{\beta}{2}\mathbb{E}_t\Vert\frac{1}{m}\sum_{i\in \left[\mathcal{m}\right]}\left(\nabla f_i(\widetilde{w}_{i}^{t}+\hat{s}_i^t) - \nabla f(\overline{w}^t)\right)\Vert^2 - \frac{\beta}{2}\mathbb{E}_t\Vert\nabla f(\overline{w}^t)\Vert^2\\
    &\leq f(\overline{w}^t) + \frac{L}{2}\mathbb{E}_t\Vert\overline{w}^{t+1}-\overline{w}^t\Vert^2 + \frac{\beta}{2m}\sum_{i\in \left[\mathcal{m}\right]}\mathbb{E}_t\Vert\nabla f_i(\widetilde{w}_{i}^{t}+\hat{s}_i^t) - \nabla f_i(\widetilde{w}_{i}^{t}) + \nabla f_i(\widetilde{w}_{i}^{t}) - \nabla f_i(\overline{w}^t)\Vert^2\\
    &\quad - \frac{\beta}{2}\mathbb{E}_t\Vert\nabla f(\overline{w}^t)\Vert^2\\
    &\leq f(\overline{w}^t) + \frac{L}{2}\mathbb{E}_t\Vert\overline{w}^{t+1}-\overline{w}^t\Vert^2 + \frac{\beta}{m}\sum_{i\in \left[\mathcal{m}\right]}\mathbb{E}_t\Vert\nabla f_i(\widetilde{w}_{i}^{t}+\hat{s}_i^t) - \nabla f_i(\widetilde{w}_{i}^{t})\Vert^2 + \frac{\beta}{m}\sum_{i\in \left[\mathcal{m}\right]}\mathbb{E}_t\Vert\nabla f_i(\widetilde{w}_{i}^{t}) - \nabla f_i(\overline{w}^t)\Vert^2\\
    &\quad - \frac{\beta}{2}\mathbb{E}_t\Vert\nabla f(\overline{w}^t)\Vert^2\\
    &\leq f(\overline{w}^t) + \frac{L}{2}\mathbb{E}_t\Vert\overline{w}^{t+1}-\overline{w}^t\Vert^2 + \frac{\beta L^2}{m}\sum_{i\in \left[\mathcal{m}\right]}\mathbb{E}_t\Vert r\frac{\mathcal{s}_i^{t}}{\Vert \mathcal{s}_i^{t}\Vert}\Vert^2 + \frac{\beta L^2}{m}\sum_{i\in \left[\mathcal{m}\right]}\mathbb{E}_t\Vert\widetilde{w}_{i}^{t} - \overline{w}^t\Vert^2 - \frac{\beta}{2}\mathbb{E}_t\Vert\nabla f(\overline{w}^t)\Vert^2\\
    &\leq f(\overline{w}^t) + \frac{L}{2}\mathbb{E}_t\Vert\overline{w}^{t+1}-\overline{w}^t\Vert^2 + \beta L^2r^2 + \frac{\beta L^2}{m}\sum_{i\in \left[\mathcal{m}\right]}\mathbb{E}_t\Vert\widetilde{w}_{i}^{t} - \overline{w}^t\Vert^2 - \frac{\beta}{2}\mathbb{E}_t\Vert\nabla f(\overline{w}^t)\Vert^2.\\
\end{align*}
We take full expectation from $0$ to $t$ on the above inequality and get:
\begin{equation}
\label{smoothness}
    \mathbb{E}\left[f(\overline{w}^{t+1})\right] \leq \mathbb{E}\left[f(\overline{w}^t)\right] + \beta L^2r^2 + \frac{L(1+2\beta L)}{2}\upsilon^t - \frac{\beta}{2}\mathbb{E}\Vert\nabla f(\overline{w}^t)\Vert^2.
\end{equation}
Equation~(\ref{smoothness}) indicates the relationship of expected global update and its norm term. Here we consider the lemma~\ref{l2} and \ref{l3},
\begin{equation}
\label{r1}
    (1 - 12\beta^2L^2)\upsilon^t \leq 30\beta^2L^2r^2 + 36\beta^2L^2\delta^{t-1} + 12\beta^2\mathbb{E}\Vert \nabla f(\overline{w}^t)\Vert^2,
\end{equation}
\begin{equation}
\label{r2}
    \delta^t \leq \frac{2m-2n}{n}\delta^{t-1} + \frac{6m-2n}{2m-n}\upsilon^t.
\end{equation}
Let formula~(\ref{r1}) multiplied by constant $p$ and formula~(\ref{r2}) multiplied by constant $q$, we take the sum of formula~(\ref{smoothness})~(\ref{r1})~(\ref{r2}),
\begin{align*}
    \mathbb{E}\left[f(\overline{w}^{t+1})\right] + p(1 - 12\beta^2L^2)\upsilon^t + q\delta^t
    &\leq \mathbb{E}\left[f(\overline{w}^t)\right] + \beta L^2r^2 + \frac{L(1+2\beta L)}{2}\upsilon^t - \frac{\beta}{2}\mathbb{E}\Vert\nabla f(\overline{w}^t)\Vert^2\\
    &\quad + 30p\beta^2L^2r^2 + 36p\beta^2L^2\delta^{t-1} + 12p\beta^2\mathbb{E}\Vert \nabla f(\overline{w}^t)\Vert^2\\
    &\quad + q\frac{2m-2n}{n}\delta^{t-1} + q\frac{6m-2n}{2m-n}\upsilon^t.\\
\end{align*}
Collecting the like term of $\upsilon^t$ and let the constants $p$ and $q$ satisfy:
\begin{equation}
\label{c1}
    p(1 - 12\beta^2L^2) = \frac{L(1+2\beta L)}{2} + q\frac{6m-2n}{2m-n}.
\end{equation}
When equation~(\ref{c1}) holds, the relationship will be simplified to:
\begin{align*}
    \mathbb{E}\left[\overline{w}^{t+1}\right] + q\delta^t
    &\leq \mathbb{E}\left[\overline{w}^t\right] + (1+30p\beta)\beta^2L^2r^2 + \left(36p\beta^2L^2+q\frac{2m-2n}{n}\right)\delta^{t-1} \\
    &\quad - \left(\frac{\beta}{2}-12p\beta^2\right)\mathbb{E}\Vert\nabla f(\overline{w}^t)\Vert^2.\\
\end{align*}
Furthermore, considering the coefficient of $\delta$ term, we let the constant $p$ and $q$ satisfy:
\begin{equation}
\label{c2}
    q = 36p\beta^2L^2+q\frac{2m-2n}{n}.
\end{equation}
According to the equation~(\ref{c1}) and (\ref{c2}), we can get the solution of $p$ and $q$ as:
\begin{align*}
    p&=\frac{L(1+2\beta L)}{2}\left[1-\left(12+\frac{36(6m-2n)n}{(2m-n)(3n-2m))}\right)\beta^2L^2\right]^{-1},\\
    q&= \frac{36n\beta^2L^2}{3n-2m}p.\\
\end{align*}
This proof requires the constants $p$ and $q$ both to be positive, thus the $\beta$ satifies $\beta\leq\frac{\sqrt{n}}{6\sqrt{6m}L}$. With this selection, the coefficient of the global gradient term maintain a positive value. Let $\zeta$ to be the coefficient of the global gradient term, we can rewrite the relationship as:
\begin{align*}
    \zeta \mathbb{E}\Vert\nabla f(\overline{w}^t)\Vert^2
    \leq \left(\mathbb{E}\left[f(\overline{w}^t)\right]+q\delta^{t-1}\right) - \left(\mathbb{E}\left[f(\overline{w}^{t+1})\right]+q\delta^{t}\right) + (1+30p\beta)\beta^2L^2r^2.\\
\end{align*}
Adding up the above formula from $0$ to $T-1$ and applying $q\leq\frac{72m\beta^2L^2}{n}$, let $\beta \leq \min\{\frac{\sqrt{n}}{6\sqrt{6m}L}, \frac{1}{2\sqrt{3}L}\}$, we have:
\begin{equation}
    \frac{1}{T}\sum_{t=1}^{T}\mathbb{E}\Vert\nabla f(\overline{w}^t)\Vert^2\leq \frac{1}{\zeta T}\left(\left(f(\overline{w}^1)-f^*\right)+\frac{72\beta^2L^2\sum_{i\in \left[\mathcal{m}\right]}\mathbb{E}\Vert w_i^0-\overline{w}^1\Vert^2}{n}\right)+(1+30p\beta)\beta^2L^2r^2.
\end{equation}
Similar to the {\ttfamily FedSAM} and {\ttfamily MoFedSAM}~\cite{FedSAM}, we select the perturbation learning rate $r=\mathcal{O}(\frac{1}{\sqrt{T}})$ that the final convergence rate approaches $\mathcal{O}(\frac{1}{T})$, which completes the proofs. Selecting some proper constants makes Theorem 4.1 hold.

\subsection{Proof of Theorem 4.4}
Based on the margin generalization bounds in~\cite{pac,generalization2}, we consider the generalization error bound on the global function $f$ as following:
\begin{equation}
\label{222}
    G_\epsilon^f\triangleq\mathbf{P}\Big(f(w+s,\varepsilon)[y]\leq \max_{j\neq y}f(w+s,\varepsilon)[j]+\epsilon\Big).
\end{equation}
$\varepsilon$ is the input data and $y$ is its ground truth. If $\epsilon=0$, we denote the equation~(\ref{222}) as $G^f$ to represent for the average misclassification rate under the global perturbation $s$. We define the $\widetilde{G}_\epsilon^f$ as the empirical estimate of the above expected margin loss trained with the heterogeneous dataset.

We assume the input data $\varepsilon$ as the normalized tensor, e.g. for an image, with the norm of $L_n$. And we consider a $L$-layer neural network with $d$ parameters per layer. The activation functions hold no biases and bounded 1-Lipschitz property. The input of each layer are bounded by a positive constant $L_w$. Under the margin value of $\epsilon$ and total $m$ data samples, with the probability of at least $1-p$, the global parameters $w$ generated by {\ttfamily FedSMOO} satisfy:
\begin{equation}
    G^f\leq \widetilde{G}_\epsilon^f + \mathcal{O}\left(\sqrt{\frac{L_l^2L_n^2d\ln(dL_l)\mathcal{V}_L+\ln\frac{L_lD}{p}}{(D-1)\epsilon^2}}\right).
\end{equation}
where $\mathcal{V}_L=\prod_{l=1}^{L}\Vert w_l\Vert^2\sum_{l=1}^L\frac{\Vert w_l\Vert_F^2}{\Vert w_l\Vert^2}$.

\begin{lemma}
\cite{pac}
    If $f(w,\varepsilon)$ is the predictor with the parameters $w$. $\mathcal{P}$ is the arbitrary distribution on $w$. We assume the data $\varepsilon$ is sampled from the $\mathbb{D}$ and its size is $D$. Thus for any $\epsilon,\delta > 0$, with the probability at least $1-p$, and the arbitrary perturbation $s$ satisfies $\mathbb{P}_s\left(\max_\varepsilon\vert f(w+s,\varepsilon)-f(w,\varepsilon)\vert_\infty\leq\frac{\epsilon}{4}\right)\geq\frac{1}{2}$, then:
    \begin{equation}
    G^f\leq\widetilde{G}_\epsilon^f+4\sqrt{\frac{KL(w+s\Vert \mathcal{P})+\log\frac{6D}{p}}{D-1}}.
    \end{equation}
\end{lemma}
This KL divergence comes from the frozen parameters $w$ and the $s$ is a random variable. It comes from the PAC-Bayes margin bounds for a linear predictors. Then we bound the inference of parameters.
\begin{lemma}
\cite{pac}
    We assume the input $\varepsilon$ is a tensor with bounded norm of $L_n$. And we consider a $L$-layer neural network with $d$ parameters per layer. The activation functions hold no biases and bounded 1-Lipschitz property. Thus, for the arbitrary input sample, let the parameters are perturbed by a bounded noise $s$ with $\Vert s_l\Vert\leq\Vert w_l\Vert$, the total change of the value of function $f$ is:
    \begin{equation}
        \Vert f(w+s,\varepsilon)-f(w,\varepsilon)\Vert\leq eL_n\prod_{l=1}^{L}\Vert w_l\Vert\sum_{l=1}^{L}\frac{\Vert s_l\Vert}{\Vert w_l\Vert}.
    \end{equation}
\end{lemma}
This lemma bounds the change from the perturbation $s$, which judges the stability of the neural network after the training. The perturbation $s$ could be an arbitrary vectors with the same size of the model parameters.

Then with the preliminary lemmas, we prove the generalization bound of our proposed method. Firstly, as the previous works~\cite{generalization2,FedSAM}, we assume the perturbation $s_l$ for each layer is a zero-mean Gaussian noise of $\mathcal{N}(0, \sigma_{l}^2)$. We consider the new parameters $w^s$ which satisfies the $\vert\Vert w_l\Vert_2-\Vert w_l^s\Vert_2\vert\leq\frac{1}{L}$, thus we have~\cite{probability}:
\begin{align*}
    \mathbb{P}\left(\prod_{l=1}^{L}\Vert w_l\Vert^\frac{1}{L}\frac{\Vert s_l\Vert}{\Vert w_l^s\Vert}>p\right)\leq 2de^{-\frac{p^2\prod_{l=1}^{L}\Vert w_l\Vert^\frac{2}{L}}{2d\sigma_l^2\Vert w_l^s\Vert^2}}.
\end{align*}
Thus, we consider the union bound over all layers by at least probability of 0.5, which is:
\begin{align*}
    \max_\epsilon\Vert f(w+s,\varepsilon)-f(w,\varepsilon)\Vert
    &\leq eL_n\prod_{l=1}^{L}\Vert w_l\Vert\sum_{l=1}^{L}\frac{\Vert s_l\Vert}{\Vert w_l\Vert}\leq e^2LL_n\prod_{l=1}^{L}\Vert w_l\Vert^\frac{L-1}{L}\sigma\sqrt{2d\ln(4dL)}.
\end{align*}
Where we select a constant variance of $\sigma=\frac{\epsilon}{4e^2LL_n\prod_{l=1}^{L}\Vert w_l\Vert^\frac{L-1}{L}\sqrt{2d\ln(4dL)}}$, we have:
\begin{align*}
    \Vert f(w+s,\varepsilon)-f(w,\varepsilon)\Vert\leq\max\Vert f(w+s,\varepsilon)-f(w,\varepsilon)\Vert
    \leq \frac{\epsilon}{4}.
\end{align*}
Thus we bound the $KL$ divergence on the distribution $\mathcal{P}$ as:
\begin{align*}
    KL(w+s\Vert \mathcal{P})
    &\leq \frac{\sum_{l=1}^L\Vert w_l\Vert_F^2}{2\sigma^2}=\frac{16e^4L^2L_n^2\prod_{l=1}^{L}\Vert w_l\Vert^\frac{2L-2}{L}d\ln(4dL)}{\epsilon^2}\sum_{l=1}^L\Vert w_l\Vert_F^2\\
    &= \frac{16e^4L^2L_n^2\prod_{l=1}^{L}\Vert w_l\Vert^2 d\ln(4dL)}{\epsilon^2}\sum_{l=1}^L\frac{\Vert w_l\Vert_F^2}{\prod_{l=1}^{L}\Vert w_l\Vert^\frac{2}{L}}\\
    &\leq \frac{16e^4L^2L_n^2\prod_{l=1}^{L}\Vert w_l\Vert^2 d\ln(4dL)}{\epsilon^2}\sum_{l=1}^L\frac{\Vert w_l\Vert_F^2}{\Vert w_l\Vert^2}.
\end{align*}
Hence, when we provide a arbitrary distribution $\mathcal{P}$, with the probability of at least $1-p$, the optimized parameters $w$ under the cover of the total dataset size $LD^\frac{1}{2L}$, we have:
\begin{equation}
    G^f\leq \widetilde{G}_\epsilon^f + \mathcal{O}\left(\sqrt{\frac{L^2L_n^2d\ln(dL)\prod_{l=1}^{L}\Vert w_l\Vert^2\sum_{l=1}^L\frac{\Vert w_l\Vert_F^2}{\Vert w_l\Vert^2}+\ln\frac{Lm}{p}}{m\epsilon^2}}\right).
\end{equation}

\end{document}